\pdfoutput=1
\documentclass[11pt]{article}

\usepackage[]{acl}

\usepackage{times}
\usepackage{latexsym}
\usepackage{graphicx}
\usepackage{booktabs}
\usepackage{subcaption}
\usepackage{adjustbox}
\usepackage{geometry}
\usepackage{enumitem}
\usepackage{multirow}
\usepackage{multicol}
\usepackage{amsmath,amsfonts,amssymb}
\usepackage{url}

\usepackage{hyperref}
\usepackage{blindtext}
\usepackage{xcolor}
\usepackage{todonotes}
\usepackage{pifont}% http://ctan.org/pkg/pifont
\newcommand{\xmark}{\ding{55}}%

\usepackage[T1]{fontenc}
\usepackage[utf8]{inputenc}

\usepackage{microtype}

\newcommand{\kwd}{KWD}
\newcommand{\tfidf}{TF-IDF}
\newcommand{\rnd}{rnd}
\newcommand{\rndF}{rnd-50}
\newcommand{\sbert}{SBERT}

\newcommand{\mOne}{Meg-1.3}
\newcommand{\mEight}{Meg-8.3}
\newcommand{\ours}{MT-NLG}
\newcommand{\Sref}[1]{\S\ref{#1}}

\newcommand{\nvidia}{$^1$}
\newcommand{\caltech}{$^2$}

\usepackage{tcolorbox}% http://ctan.org/pkg/tcolorbox
\definecolor{lightBlue}{rgb}{0.8, 0.9, 1.0}% Rule colour
\definecolor{lightRed}{rgb}{1.0, 0.90, 0.90}% Rule colour

\newtcbox{\bluebox}{on line, box align=base, colback=lightBlue,colframe=white,size=fbox,arc=3pt, before upper=\strut, top=-2pt, bottom=-4pt, left=-2pt, right=-2pt, boxrule=0pt}

\newtcbox{\redbox}{on line, box align=base, colback=lightRed,colframe=white,size=fbox,arc=3pt, before upper=\strut, top=-2pt, bottom=-4pt, left=-2pt, right=-2pt, boxrule=0pt}

\newcommand{\dashifted}{\raisebox{0.5\depth}{\tiny$\downarrow$}}
\newcommand{\upshifted}{\raisebox{0.5\depth}{\tiny$\uparrow$}}

\newcommand{\dar}[1]{{\scriptsize\redbox{\dashifted{#1}}}}

\newcommand{\uab}[1]{{\scriptsize\bluebox{\upshifted{#1}}}}

\title{Few-shot Instruction Prompts for Pretrained Language Models to Detect Social Biases}

\author{Shrimai Prabhumoye\nvidia{}, Rafal Kocielnik\caltech{}, Mohammad Shoeybi\nvidia{}, \\
\bf Anima Anandkumar\nvidia{}$^,$\caltech{}, Bryan Catanzaro\nvidia{} \\
  \nvidia{}NVIDIA, \caltech{}California Institute of Technology \\
  \texttt{\{sprabhumoye@nvidia.com, rafalko@caltech.edu\}} \\}

\begin{document}
\maketitle
\begin{abstract}
\textcolor{purple}{\textbf{\textit{Warning:}} this paper contains content that may be offensive or upsetting.}

Detecting social bias in text is challenging due to nuance, subjectivity, and difficulty in obtaining good quality labeled datasets at scale, especially given the evolving nature of social biases and society. 
To address these challenges, we propose a few-shot instruction-based method for prompting pre-trained language models (LMs).
We select a few class-balanced exemplars from a small support repository that are closest to the query to be labeled in the embedding space.
%The novelty of our method relies on selecting a few label-balanced exemplars from a small support repository that are closest to the post to be labeled in the embedding space.
%Instead of using all the given labeled samples for prompting, the novelty of our method relies on selecting a few label-balanced exemplars based on similarity in an embedding space.
We then provide the LM with instruction that consists of this subset of labeled exemplars, the query text to be classified, a definition of bias, and prompt it to make a decision.
%We demonstrate that LMs are capable of detecting different types of fine-grained biases without finetuning for these tasks.
We demonstrate that large LMs used in a few-shot context can detect different types of fine-grained biases with similar and sometimes superior accuracy to fine-tuned models.
We observe that the largest 530B parameter model is significantly more effective in detecting social bias compared to smaller models (achieving at least $13$\% improvement in AUC metric compared to other models).
It also maintains a high AUC (dropping less than $2$\%) when the labeled repository is reduced to as few as $100$ samples. 
Large pretrained language models thus make it easier and quicker to build new bias detectors.
%It drops less than $5$\% AUC points when the size of labeled repository is as low as $100$ samples as opposed to a $9.8$\% AUC point drop for the 1.3B parameter model).
% Recent success in large pre-trained language models (LMs) has driven the field to leverage the inherent knowledge present in them by prompting.
% This paper explores the ability of LMs to detect social bias in text especially when they have access to limited labeled examples.
% We provide the LM with a few labeled examples of bias, the unlabeled text to be classified, a definition of bias, and prompt it to make a decision.
% We demonstrate the effectiveness of our approach by presenting benchmark results on $8$ bias detection tasks spanning across two datasets.
% We showcase that LMs are capable of detecting different types of biases without finetuning or optimizing for multiple tasks.
% We illustrate this through comprehensive studies with small sizes of labeled repositories, ablation studies showing the contribution of different semantic components of our approach, robustness of LMs against perturbations in labels, and an extensive qualitative analysis of the LM predictions.
% [For largest model, we increase by X]
\end{abstract}

\section{Introduction}

% \begin{itemize}
%     \item Success of pre-trained LMs on a set of tasks - commonsense, language understanding, qa etc
%     \item In this work, we explore the ability of the LMs to diagnose bias in text. Explain social bias in text and how it is studied, what are the challenges in studying it.
%     \item Current approaches rely on finetuning (high cost) and large labeled data (not easily available and costly to collect). finetuning requires optimization for multi-task performance
%     \item We address these issues by proposing leveraging large LMs to / proposing to prompt LMs . LMs already have knowledge and we use to to identify bias. How do we do this? We give text and then instruction and ... This does not require finetuning or large labeled dataset or any additional optimization for different tasks.
%     \item In this paper, we show results on two datasets, binary and multi-class classification, comprehensive ablations to understand contribution of each component, experiments with reduced datasets, and robustness of the model for wrong/flipped labels.
% \end{itemize}
Detecting social bias in text is of utmost importance as stereotypes and biases can be projected through language~\cite{fiske1993controlling}.
Detecting bias is challenging because it can be expressed through seemingly innocuous statements which are implied and rarely explicit, and the interpretation of bias can be subjective leading to noise in labels. 
In this work, we focus on detecting social bias in text as defined in \citet{sap-etal-2020-social} using few-shot instruction-based prompting of pre-trained language models (LMs).
%we address these challenges by proposing a novel approach to selection of exemplars in few-shot instruction-based prompting using pre-trained language models (LMs).

Current approaches that detect bias require large labeled datasets to train the models~\cite{chung-etal-2019-conan,waseem-hovy-2016-hateful,zampieri-etal-2019-predicting,davidson2017automated}.
Collecting such labeled sets is an expensive process and hence they are not easily available.
Furthermore, most of the prior work relies on finetuning~\cite{sap-etal-2020-social,mandl-etal-2019-hasoc,zampieri-etal-2019-predicting} neural architectures which is costly in case of large LMs~\cite{strubell2019energy} and access to finetune large LMs may be limited~\cite{brown-etal-2020-gpt3}.
Prior work on bias detection has not focused on modeling multiple types of biases across datasets as it requires careful optimization to succeed~\cite{hashimoto2017joint,sogaard-goldberg-2016-deep,ruder2017overview}.
%Another limitation of finetuning approaches is that it requires careful optimization to succeed on multiple tasks or on a variety of datasets~\cite{hashimoto2017joint,sogaard-goldberg-2016-deep,ruder2017overview}.
Finetuning a model can also lead to over-fitting especially in case of smaller train sets and to catastrophic forgetting of knowledge present in the pre-trained model \cite{fatemi2021improving}.
Moreover, finetuning approaches are prone to be affected by noisy labels~\cite{song2022learning} which is especially an issue with datasets for bias detection.
The human labeling used to annotate these datasets can introduce bias and noisy labels~\cite{hovy2021five}.

\begin{figure*}[!t]
\centering
{
\includegraphics[width=\linewidth]{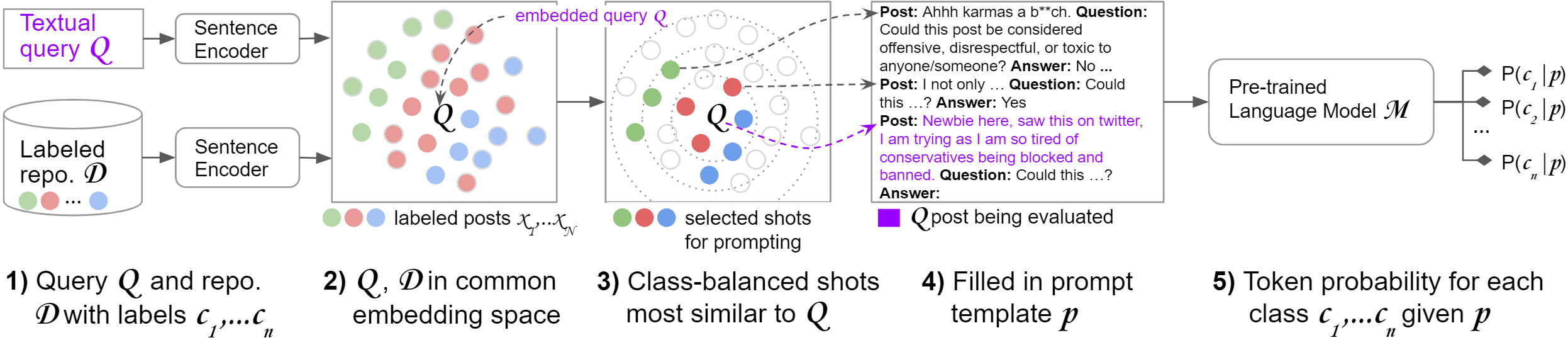}
}
\vspace{-1.0em}
\caption{Overview of our approach: We use a sentence encoder to project the query $\mathbf{Q}$ and the textual posts $\mathbf{x}_1,\ldots \mathbf{x}_N$ from labeled repository $\mathcal{D}$ to the same embedding space. We use cosine similarity metric to select equal number of posts with highest similarity to $\mathbf{Q}$ from each class $\mathbf{c}_1\dots \mathbf{c}_n$ as shots. The tokens of the selected shots and their labels are concatenated along with the definition of bias and query to fill-in instruction template $\mathbf{p}$, and finally passed to the pre-trained LM to make a prediction based on conditional token probability for each class.}
\vspace{-1.0em}
\label{fig:overview}
\end{figure*}

We harness the knowledge present in large scale pre-trained language models~\cite{davison-etal-2019-commonsense,zhou-etal-2020-evaluating,petroni-etal-2019-language,zhong-etal-2021-factual,shin-etal-2020-autoprompt} to detect a rich set of biases.
%Pre-trained LMs are known to possess commonsense knowledge~\cite{davison-etal-2019-commonsense,zhou-etal-2020-evaluating}, and have been successfully prompted for cloze questions tasks~\cite{petroni-etal-2019-language,zhong-etal-2021-factual,shin-etal-2020-autoprompt}.
%we prompt the LM to utilize the knowledge it contains to detect bias.
Our method prompts the LM with a textual post and labeled exemplars along with instructions to detect bias in the given post.
We explore the capabilities of LMs to flexibly accommodate different dimensions of bias without any finetuning and with limited access to labeled samples (few-shot classification).
% In Figure \ref{fig:instruction_formulation}, we see that $\mathbf{X}$ is provided as context and $\mathbf{Y}$ is provided as an instruction.
% In Figure \ref{fig:post_examples} we can see that bias is present in implicit form and most heuristic based techniques which rely on surface forms are unable to detect implied consequences of seemingly innocuous statements.

Prompt-engineering plays a central role in finetuning-free approaches ~\cite{liu2021pre}.
It is the process of creating a prompting function that results in the best performance on the desired downstream task. 
Prompt-engineering can be performed by a human engineer who manually creates the desired prompts using domain expertise and intuition. 
It can also be performed by sophisticated algorithms that search for the best template for the downstream task but this too requires learning of a small number of weights (finetuning).

\textbf{Our approach:} 
We provide the LM with exemplars where semantically similar text is used in both biased and unbiased contexts.
%The motivation for our approach is to provide the LM with exemplars where semantically similar text is used in both biased and unbiased contexts.
This specifically can be useful in identifying implicit biases.
To achieve that, we use prompt-engineering approach in combination with a novel method to sample class-balanced exemplars in case of few-shot bias classification.
%Prior work sample exemplars using uniform~\cite{gao2020making,logan2021cutting} or random~\cite{brown-etal-2020-gpt3} sampling from a labeled repository.
We propose the use of sentence similarity metric to sample exemplars instead of uniform~\cite{gao2020making,logan2021cutting} or random~\cite{brown-etal-2020-gpt3} sampling explored before.
As shown in Figure~\ref{fig:overview}, we first utilize sentence encoder to project the labeled posts and the query post to the same embedding space.
We use a similarity metric to identify posts from labeled repository that are closest in meaning to the query post.
We then select an equal number of exemplars from each class label to be provided as context for few-shot classification.
To summarize, our contributions are as follows:
\begin{itemize}[leftmargin=*]
    \item To our best knowledge, we are the first to adopt few-shot instruction based techniques to detect social bias without finetuning.
    %We show how the knowledge of pre-trained LMs can be used to detect different types of biases in text.
    %We adopt few-shot instruction based techniques to detect bias without any finetuning.
    \item We propose a novel approach to select class-balanced exemplars for few-shot classification (\Sref{sec:method}) %using semantic similarity sampling (\Sref{sec:method}).
    \item We establish few-shot based benchmarks on eight binary and multi-class classification tasks across two datasets, even beating the fine-tuning techniques on three tasks (\Sref{sec:results}).
    %We establish benchmark performance of LMs on six bias binary classification tasks and two multi-class classification tasks covering two separate datasets (\Sref{sec:results}).
    %We observe that our best performing model performs better than finetuned state-of-the-art approach on two binary classification tasks.
    %These experiments show the flexibility of our technique in using different bias definitions.
    \item We demonstrate that our technique maintains performance with smaller repository sizes (less than $2$\% AUC point drop in downsizing the labeled repository from 35k to 100 samples)(\Sref{sec:analysis}). 
    %and to perturbations in labels (only 0.99 F1 point drop when half of the shot labels are randomized) (\Sref{sec:analysis}).
    %Furthermore, we show multiple comprehensive analyses such as experiments with smaller sizes of labeled repositories, ablation studies showing the contribution of different semantic components of our technique, robustness of LMs against perturbations in labels, and an exhaustive qualitative analysis of the LM predictions (\Sref{sec:analysis}).
    \item Finally, we scale our technique to a large LM with 530B parameters and illustrate that it can achieve at least $13$\% AUC improvement compared to other models on majority of tasks.
\end{itemize}

Our proposed technique does not require any additional complex tuning to perform multiple tasks and is flexible to identify a diverse set of biases focusing on coarse-grained (binary classification) as well as fine-grained (if the text was targeted or untargeted insult, who is targeted in the text, etc) tasks.
Our experiments show that pretrained language models are robust against noisy labels.
We demonstrate that the models are able to predict the correct label more than a third of the time even when provided with $100$\% flipped labels.
Additionally, we present ablations to understand the contribution of different semantic components of our method, and an exhaustive qualitative analysis of the LM predictions.

\section{Methodology}
\label{sec:method}
We study the ability of pre-trained LMs to detect implicit bias in text.
We also investigate if large LMs are capable of doing so with limited access to labeled samples (few-shot classification) and without any finetuning.
We propose to sample class-balanced exemplars from a labeled repository based on their semantic similarity with the query post.
We only provide the LM with a few examples of bias, the textual post to be classified and a definition of bias and prompt it to make a decision.
We present all our exemplars as Question-Answer pairs in context.
Our approach consists of syntactic structured component and a semantic content component.
The syntactic structure is provided by special tags such as \textit{Post}, \textit{Question} and \textit{Answer}.
The semantic content consists of the textual exemplars chosen from the labeled repository and their labels.

Formally, we have a textual post $\mathbf{Q}$ and a definition of bias $\mathbf{d}$.
We want to categorize text $\mathbf{Q}$ in $\mathbf{C}$ classes. 
We provide a language model $\mathcal{M}$ with $\mathbf{Q}$ and $\mathbf{d}$, and we check the probability $p_{\mathcal{M}}(\mathbf{c}_i | \mathbf{Q}; \mathbf{d})$ of each class $\mathbf{c}_i \in \mathbf{C}$, where $;$ denotes concatenation of the two strings $\mathbf{Q}$ and $\mathbf{d}$.
We consider the class with the highest probability to be the prediction of $\mathcal{M}$ for the given $\mathbf{Q}$ and $\mathbf{d}$.

For example as shown in Figure~\ref{fig:overview}, consider the binary classification task of predicting if a piece of text is offensive or not.
Here, the text $\mathbf{Q}$ to be labeled is \textit{Newbie here, saw this on twitter, I am trying as I am so tired of conservatives being blocked and banned.} and the definition $\mathbf{d}$ of offensiveness used to annotate the text is \textit{Could this post be considered offensive, disrespectful, or toxic to anyone/someone?}.
The final input to model $\mathcal{M}$ would be $\mathbf{p} = [``\mathtt{Post:}"; \mathbf{Q}; ``\mathtt{Question:}"; \mathbf{d}; ``\mathtt{Answer:}"]$, where we concatenate the tags \textit{Post}, \textit{Question} and \textit{Answer}.
Since it is a binary classification task, we calculate the probabilities of the tokens \textit{Yes} and \textit{No} in the following manner: $p_{\mathcal{M}}(\mathtt{``Yes"} | \mathbf{p})$ and $p_{\mathcal{M}}(\mathtt{``No"} | \mathbf{p})$.
The token that has the higher probability is considered the prediction of the task.

Similarly, in case of a multi-class classification task of predicting if the text contains \textit{hate}, \textit{offense} or \textit{profanity}, the $\mathbf{d}$ is \textit{What does this post contain - hate, offense, profanity?}
We calculate the probability of the tokens \textit{hate}, \textit{offense} and \textit{profanity} given the input $\mathbf{p}$.

In the zero-shot case, only the input $\mathbf{p}$ as described above is provided as input to $\mathcal{M}$.
In the few-shot case, we need a labeled repository $\mathcal{D}$ where each sample $(\mathbf{x}_i, \mathbf{c}_i)$ is a tuple of the textual post and the class of the post respectively.
We also know the definition $\mathbf{d}$ to be used for classification.
In the $k$-shot case, $k$ samples are chosen from $\mathcal{D}$.
The input $\mathbf{p}$ to the model $\mathcal{M}$ in this case is the concatenation of the following strings $\mathbf{p} = [\mathbf{x}_1; \mathbf{d}; \mathbf{c}_1; \ldots; \mathbf{x}_k; \mathbf{d}; \mathbf{c}_k; \mathbf{Q}; \mathbf{d}]$.
We add the tags \textit{Post}, \textit{Question} and \textit{Answer} for structure.

\paragraph{Selection of k-shots}
In case of few-shot classification, we select $k$ exemplars from $\mathcal{D}$ repository to be provided as context to $\mathcal{M}$.
Instead of randomly selecting the $k$ exemplars, we select the samples that are closest in meaning to the text $\mathbf{Q}$ which we want to classify.
We project $\mathbf{Q}$ and all the text samples from $\mathcal{D}$ in the same embedding space.
We use cosine similarity and select the $k$ exemplars that have the highest cosine similarity scores.
We also have an additional constraint of selecting equal number of exemplars from each class $\mathbf{C}$ to ensure balanced representation of labels.
For example, in case of $32$-shot binary classification, we select $16$ positive exemplars and $16$ negative exemplars that are closest in meaning to $\mathbf{Q}$. 

\section{Experiments and Results}
\label{sec:experiments}

\begin{table*}[t]
\centering
\small{
\begin{tabular}{@{}l l r r @{\hskip 0.15in} r r @{\hskip 0.15in} r r @{\hskip 0.15in} r r@{}}
\textbf{Model} & \textbf{Sampling} & \multicolumn{2}{c}{\textbf{Offensive}}{\hskip 0.3in} & \multicolumn{2}{c}{\textbf{Intent}}{\hskip 0.3in} & \multicolumn{2}{c}{\textbf{Lewd}}{\hskip 0.3in}  & \multicolumn{2}{c}{\textbf{Group}}\\
\toprule
& & AUC & F1 & AUC & F1 & AUC & F1 & AUC & F1  \\
\cmidrule{3-10}
SC & & - & 78.80 & - & 78.60 & - & 80.70 & - & 69.90 \\
\cmidrule{3-10}
\kwd & - & 58.31 & 70.94 & 56.94 & 67.17 & 59.83 & 35.39 & 55.50 & 57.16 \\
\tfidf & \tfidf & 63.64 & 72.76 & 65.00 & 69.71 & 56.77 & 21.67 & 65.57 & 59.74 \\
\midrule
\mOne & \rnd & 55.02 & 64.45 & 55.25 & 51.57 & 49.80 & 0.00 & 52.33 & 28.73 \\
\mOne & \rndF & \uab{4.69\%} 57.60 & 64.13 & \uab{5.05\%} 58.04 & 57.79 & \uab{10.20\%} 54.88 & 19.02 & \uab{6.34\%} 55.65 & 52.20 \\
\vspace{5pt}
\mOne & \tfidf & \uab{12.92\%} 62.13 & 73.49 & \uab{14.39\%} 63.20 & 69.04 & \uab{23.21\%} 61.36 & 22.11 & \uab{16.87\%} 61.16 & 60.61 \\

\mEight & \rnd & 61.49 & 74.11 & 61.96 & 68.17 & 50.10 & 0.86 & 60.68 & 48.40 \\
\mEight & \rndF & \uab{5.94\%} 65.14 & 76.20 & \uab{4.71\%} 64.88 & 73.05 & \uab{27.09\%} 63.67 & 24.10 & \uab{4.55\%} 63.44 & 63.01 \\
\vspace{5pt}
\mEight & \tfidf & \uab{7.38\%} 66.03 & 77.53 & \uab{7.31\%} 66.49 & 74.39 & \uab{38.26\%} 69.27 & 28.00 & \uab{7.79\%} 65.41 & 65.31 \\

\ours & \rnd & 75.77 & 79.76 & 72.97 & 75.24 & 51.80 & 7.47 & 71.56 & 64.92 \\
\ours & \rndF & \uab{0.98\%} 76.51 & 79.49 & \uab{3.65\%} 75.63 & 78.38 & \uab{41.58\%} 73.34 & 32.68 & \uab{4.47\%} 74.76 & 71.69 \\
\ours & \tfidf & \uab{3.73\%} \textbf{78.60} & \textbf{82.19} & \uab{4.99\%} \textbf{76.61} & \textbf{79.77} & \uab{51.62\%} \textbf{78.54} & \textbf{41.06} & \uab{7.22\%} \textbf{76.73} & \textbf{73.74} \\
\bottomrule
\end{tabular}
}
\vspace{-0.5em}
\caption{Results for the $32$-shot prompting on four binary classification tasks offensive, intent, lewd and group from SBIC dataset (\Sref{sec:datasets}). The best performance in each task is presented in bold. We show the relative percentage improvement (\uab{}) in AUC score compared to the \rnd{} sampling. The improvement gained by \ours{} by using \tfidf{} vs. \rndF{} is less compared to smaller models.}
\vspace{-1.0em}
\label{tab:sbic_results}
\end{table*}

\subsection{Datasets}
\label{sec:datasets}

We consider two separate datasets and a total of eight bias classification tasks.
Note that models finetuned on one dataset would need to be further optimized or finetuned on the other dataset but this is not the case for our approach.

\paragraph{Social Bias Frames (SBIC)} This dataset~\cite{sap-etal-2020-social} contains fine-grained categorization of textual comments to better model the pragmatic frames in which people project social biases and stereotypes onto others.
It contains four binary classification tasks and one multi-class classification task (\%age positive samples in test set are shown in brackets): 
(1) \textbf{\textit{offensive}} task (57.8\% pos): predict if the text is offensive or not, 
(2) \textbf{\textit{intent}} task (53.1\% pos): predict if the text is an intentional insult or not, 
(3) \textbf{\textit{lewd}} task  (9.6\% pos): predict if the text contains lewd language or not, 
(4) \textbf{\textit{group}} task  (41.1 \% pos): predict if the text is offensive to a group or an individual, and 
(5) \textbf{\textit{target group (WHO)}} task: if the text is offensive to a group then identify the group targeted in the text.
We design the target group identification as a seven-way classification task where the target group categories are - \textit{body, culture, disabled, gender, race, social, victim}.

\citet{sap-etal-2020-social} treat these five tasks as a single generative task where the entire frame is generated token by token.
We treat them as five separate classification tasks.

\paragraph{HASOC} This dataset~\cite{mandl-etal-2019-hasoc} is released in three Indo-European Languages. 
We only focus on English tasks.
It consists of two binary classification tasks and one multi-class classification task: 
(1) \textbf{\emph{HOF}} task (25.0\% pos): is a coarse-grained task of determining whether a post contains hate, offensive, and profane content (as one label) or not, 
(2) \textbf{\emph{HOP}} task: is a fine-grained task that considers the type of offense. 
This is a three-way classification task to predict if a post contains hate speech, offensive language or profane content.
(3) \textbf{\emph{Target}} task (85.1\% pos): entails further categorizing the text as targeted or untargeted insult.
Only posts that have a positive label in \emph{HOF} task are considered for \emph{HOP} and \emph{Target} tasks.

\subsection{Baselines}
\label{sec:baselines}
We consider two simple heuristics which can prove to be strong baselines for these tasks due to high correlation of certain keywords with labels.

\paragraph{Keyword-based (\kwd)}
We use 3 common keyword-based baselines . 
The \emph{LDNOOBW} dataset \cite{GitHubLD98:online} contains 403 banned English words and has been used in prior research~\cite{salminen2019online,AIandthe49:online}.
%and by organizations such as \emph{OpenStreetMap}, \emph{Google}, and \emph{Stack Overflow} \cite{AIandthe49:online}. 
The dataset of \emph{bad, offensive and profane words} 
%developed by CMU's Luis von Ahn's group 
\cite{UsefulRe32:online} contains more than 1300 English terms that could be found offensive or profane. 
Finally, \emph{obscenity and profanity} dataset contains more than 1600 popular English keywords for profanities and their variations grouped into 10 categories including \textit{sexual acts}, \textit{sexual orientation}, \textit{racial/ethnic slurs}, \textit{religious offense} \cite{GitHubsu77:online}.

We assign a positive label to a post if it contains at least one keyword from the list in a given dataset. 
As these datasets don't perfectly align with the categories from \cite{sap-etal-2020-social} and \cite{mandl-etal-2019-hasoc}, in Table~\ref{tab:sbic_results} and~\ref{tab:hasoc_results} we report the metrics for the best performing dataset (\kwd). 

\paragraph{\tfidf{}} We use Term Frequency-Inverse Document Frequency (\tfidf{}) to project the text $\mathbf{Q}$ and the posts from $\mathcal{D}$ in the common embedding space \cite{TfIdfSciKit:online}.
%\footnote{\url{https://scikit-learn.org/stable/modules/generated/sklearn.feature_extraction.text.TfidfVectorizer.html}}
For $k$-shot classification, we select $k/|\mathbf{C}|$ posts with the highest cosine similarity score from each class i.e we select equal number of posts from each class.
We then average the similarity scores for the selected samples from each class.
The class that has the highest average score is considered the prediction of the \tfidf{} baseline.

\subsection{Sampling Techniques}

\paragraph{Random (\rnd)} 
$k$ exemplars are randomly selected from repository $\mathcal{D}$ are provided as context to the language models.
This sampling is agnostic to the labels of the exemplars selected.

\paragraph{Class Balanced Random (\rndF)} 
In this technique, we randomly select $k$ class balanced exemplars from $\mathcal{D}$ i.e. in case of binary classification, we ensure that $k/2$ exemplars are randomly selected from the positive class and $k/2$ exemplars are randomly selected from the negative class.

\paragraph{Similarity Based}
This technique selects $k$ exemplars based on their semantic similarity to the query to be classified $\mathbf{Q}$.
As described in Section \Sref{sec:baselines}, we use \tfidf{} representation to encode $\mathbf{Q}$ and exemplar text in $\mathcal{D}$.
We select $k/|\mathbf{C}|$ exemplars with the highest cosine similarity score from each class.

\subsection{Evaluation Metric}
\label{sec:eval_metric}

Following~\citet{sap-etal-2020-social}, we use binary F1 score of the positive class for measuring the performance of our models on offensive, intent, lewd, and group tasks.
Additionally, we also report area under the curve (AUC) which measures the ability of a classifier to distinguish between classes \cite{AUCSciKit:online}.
For the WHO task we report the weighted F1 scores and AUC.
Similar to~\citet{mandl-etal-2019-hasoc}, we report F1-macro (F1m) and F1 weighted (F1w) for HOF, HOP and Target tasks.
%\footnote{\url{https://scikit-learn.org/stable/modules/model_evaluation.html}}

\subsection{Modeling Details}
\label{sec:model_details}

\begin{table}[t]
\centering
\small{
\begin{tabular}{@{}l l r r @{\hskip 0.3in} r r @{}}
\textbf{Model} & \textbf{Sam} & \multicolumn{2}{c}{\textbf{HOF}}{\hskip 0.3in} & \multicolumn{2}{c}{\textbf{Target}}\\
\toprule
 & & F1m & F1w & F1m & F1w \\
\cmidrule{3-6}
HS & & 78.82 & 83.95 & 51.11 & 75.63 \\
\cmidrule{3-6}
\kwd & & 56.88 & 71.68 & 41.31 & 53.17 \\
\tfidf & \tfidf & 51.71 & 60.45 & 45.49 & 63.02 \\
\mOne & \rndF & 52.06 & 61.51 & 40.63 & 50.17 \\
\mOne & \tfidf & 52.96 & 60.88 & 45.36 & 57.33 \\
\mEight & \rndF & 58.64 & 65.27 & 45.62 & 59.79 \\
\mEight & \tfidf & 58.52 & 63.99 & \textbf{51.25} & \textbf{67.16} \\
\ours & \rndF & 63.02 & 74.19 & 30.02 & 32.57 \\
\ours & \tfidf & \textbf{65.81} & \textbf{75.22} & 36.27 & 42.48 \\
\bottomrule
\end{tabular}
}
\vspace{-0.5em}
\caption{Results from the $32$-shot prompting on two HASOC binary classification tasks (\Sref{sec:datasets}). The best performance in each task is presented in bold.}
\vspace{-1.0em}
\label{tab:hasoc_results}
\end{table}

For language model $\mathcal{M}$, we use off-the-shelf pre-trained models.
%We use BERT large uncased (\textbf{\bert{}})~\cite{devlin-etal-2019-bert} model (336M parameters), GPT-2 medium (\textbf{\gpt{}})~\cite{radford2019language} model (345M parameters), and \textbf{\mOne} model which is a Megatron 1.3B parameter model pre-trained using the toolkit in \citet{shoeybi2019megatron}.
We use Megatron 1.3B parameter model (\textbf{\mOne}) and  Megatron 8.3B parameter model (\textbf{\mEight{}}) models pre-trained using the toolkit in \citet{shoeybi2019megatron}.
To understand the scaling of our technique to larger LMs, we perform experiments with \textbf{\ours{}} which is a GPT-style 530B parameter model~\cite{Patwary2021using}.

We use the train sets of SBIC and HASOC as labeled repository $\mathcal{D}$ for sampling exemplars for $k$-shot classification.
We pre-process the SBIC train set to ensure that the test set does not overlap with the train set.
We compute levenshtein distance\footnote{\url{https://pypi.org/project/python-Levenshtein/}} $l$ between each sentence $\mathbf{Q}$ in the test set and labeled repository $\mathbf{x}_i$.
If the ratio $r = \frac{2*l}{|\mathbf{Q}| * |\mathbf{x}_i|}$, is less than $0.1$, then we discard the train sentence $\mathbf{x}_i$.
Here $|.|$ indicates the length of the sentence in terms of number of characters.
The ratio $r$ tells us if the train sentence $\mathbf{x}_i$ can be transformed to test sentence $\mathbf{Q}$ by changing less than $10$\% of the characters.

We use \tfidf{} as a baseline.
The $k$ shots picked by \tfidf{} are also used in experiments with LMs in $k$-shot classification.
All the binary classification tasks are performed with $k=32$ for uniformity.
The bias definitions used for all the tasks are provided in Appendix~\ref{sec:bias_def}.

\begin{table}[t]
\centering
\small{
\begin{tabular}{@{}l l r r r r@{}}
\textbf{Model} & \textbf{Sam} & \multicolumn{2}{c}{\textbf{WHO (7-way)}} & \multicolumn{2}{c}{\textbf{HOP (3-way)}}\\
\toprule
 & & AUC & F1w & F1m & F1w \\
\cmidrule{3-6}
HS & & - & - & 54.46 & 72.77 \\
\cmidrule{3-6}
\tfidf & \tfidf & 72.26 & 42.94 & 32.90 & 32.46 \\
\mOne & \rndF & 63.40 & 28.87 & 34.41 & 38.68 \\
\mOne & \tfidf & 67.29 & 32.84 & 35.11 & 38.29 \\
\mEight & \rndF & 76.08 & 43.25 & 28.64 & 29.23 \\
\mEight & \tfidf & 82.27 & 51.21 & 25.24 & 25.05 \\
\ours & \rndF & 86.67 & 64.82 & \textbf{48.02} & \textbf{51.54} \\
\ours & \tfidf & \textbf{88.60} & \textbf{67.91} & 46.28 & 48.10 \\
\bottomrule
\end{tabular}
}
\vspace{-0.5em}
\caption{Results for the multi-class classification tasks from HASOC and SBIC (\Sref{sec:datasets}). Due to the number of classes, the \textit{WHO} classification is performed with $k=28$ and \textit{HOP} is performed with $k=3$. The best performance in each task is presented in bold.}
\vspace{-1.0em}
\label{tab:multi_class_results}
\end{table}

\subsection{Results}
\label{sec:results}

\begin{table*}[t]
\centering
\small{
\begin{tabular}{@{}l c r r @{\hskip 0.30in} c r r r@{}}
\textbf{Model} & \multicolumn{1}{c}{\textbf{Sampling}} & \multicolumn{1}{c}{\textbf{AUC}} & \multicolumn{1}{c}{\textbf{F1}}{\hskip 0.4in}  & \multicolumn{1}{c}{\textbf{Sampling}} & \multicolumn{1}{c}{\textbf{AUC}} & \multicolumn{1}{c}{\textbf{F1}}  &
\textbf{$\mathcal{D}$ Size} \\
\toprule
% \sbert & 61.15 $\pm$ 0.00 & 69.92 $\pm$ 0.00 & 35k \\
% \sbert & \textbf{64.86} $\pm$ 1.62 & 69.94 $\pm$ 0.98 & 10k \\
% \sbert & 64.78 $\pm$ 1.93 & \textbf{74.21} $\pm$ 0.56 & 1k \\
% \sbert & 62.01 $\pm$ 1.61 & 72.98 $\pm$ 1.09 & 100\\
% & \\
\tfidf & \tfidf & 62.74 $\pm$ 0.00 & 55.97 $\pm$ 0.00 & - & - & - & 35k \\
\tfidf & \tfidf & \dar{1.56\%} 61.76 $\pm$ 0.21 & 55.43 $\pm$ 0.23 & - & - & - & 10k \\
\tfidf & \tfidf & \dar{9.26\%} 56.93 $\pm$ 0.72 & 51.39 $\pm$ 0.37 & - & - & - & 1k \\
\vspace{5pt}
\tfidf & \tfidf & \dar{10.60\%} 56.09 $\pm$ 1.19 & 50.38 $\pm$ 0.53 & - & - & - & 100 \\

\mOne & \tfidf & 61.96 $\pm$ 0.00 & 56.31 $\pm$ 0.00 &  \rndF & 56.54 $\pm$ 0.00 & 48.29 $\pm$ 0.00 & 35k \\
\mOne & \tfidf & \dar{0.06\%} 61.92 $\pm$ 0.30 & 56.34 $\pm$ 0.17 & \rndF &  56.55 $\pm$ 0.54 & 48.07 $\pm$ 0.53 & 10k \\
\mOne & \tfidf & \dar{3.20\%} 59.98 $\pm$ 0.13 & 53.61 $\pm$ 0.10 & \rndF & 57.00 $\pm$ 0.07 & 48.40 $\pm$ 0.05 & 1k \\
\vspace{5pt}
\mOne & \tfidf & \dar{5.47\%} 58.57 $\pm$ 0.38 & 50.30 $\pm$ 0.20 & \rndF & 56.99 $\pm$ 0.51 & 47.93 $\pm$ 1.05 & 100 \\

\mEight & \tfidf & 66.80 $\pm$ 0.00 & 61.31 $\pm$ 0.00 & \rndF & 64.28 $\pm$ 0.00 & 59.09 $\pm$ 0.00 & 35k \\
\mEight & \tfidf & \uab{1.03\%} 67.49 $\pm$ 0.01 & 63.20 $\pm$ 0.04 & \rndF & 64.02 $\pm$ 0.04 & 58.88 $\pm$ 0.12 & 10k \\
\mEight & \tfidf & \dar{1.60\%} 65.73 $\pm$ 0.22 & 60.61 $\pm$ 0.19 & \rndF & 64.93 $\pm$ 0.45 & 59.50 $\pm$ 0.36 & 1k \\
\vspace{5pt}
\mEight & \tfidf & \dar{2.93\%} 64.84 $\pm$ 0.41 & 59.92 $\pm$ 0.24 & \rndF & 64.73 $\pm$ 0.51 & 59.50 $\pm$ 0.29 & 100 \\

\ours & \tfidf & 77.62 $\pm$ 0.00 & 69.19 $\pm$ 0.00 & \rndF & 75.06 $\pm$ 0.00 & 65.56 $\pm$ 0.00 & 35k \\
\ours & \tfidf & \uab{0.0\%} 77.62 $\pm$ 0.10 & 69.24 $\pm$ 0.20 & \rndF & 75.31 $\pm$ 0.28 & 65.82 $\pm$ 0.20 & 10k \\
\ours & \tfidf & \dar{0.95\%} 76.88 $\pm$ 0.31 & 67.47 $\pm$ 0.28 & \rndF & 75.30 $\pm$ 0.46 & 65.86 $\pm$ 0.32 & 1k \\
\ours & \tfidf & \dar{1.84\%} 76.19 $\pm$ 0.28 & 66.71 $\pm$ 0.36 & \rndF & 75.62 $\pm$ 0.71 & 66.11 $\pm$ 0.47 & 100 \\
\bottomrule
\end{tabular}
}
\vspace{-0.5em}
\caption{Results of the $32$-shot prompting on the four SBIC classification tasks with decreasing sizes of labeled repository $\mathcal{D}$. We show the \textit{std} for \textit{AUC} and \textit{F1} metric on 3 versions of the dataset downsized with different seeds. We show the relative percentage improvement (\uab{}) or decrements (\dar) in AUC score compared to the $35$k support repository size. The largest model \textit{\ours} experiences the smallest decrease in performance (only $1.84$\% AUC).}
\vspace{-1.0em}
\label{tab:train_set_results}
\end{table*}

The main results for the six binary classification tasks in the $32$-shot case are shown in Tables~\ref{tab:sbic_results} and~\ref{tab:hasoc_results}.\footnote{In \Sref{sec:sbert} we show additional analysis with more sampling techniques which can perform better for these bias tasks. But we show in Table~\ref{tab:sbert_res} and Table~\ref{tab:data_strat} that these sampling techniques take advantage of the keywords that are used only in one specific contexts. Hence, they cannot not generalize to other tasks.}
We show the results from \citet{sap-etal-2020-social} (\textbf{SC}) and \citet{mandl-etal-2019-hasoc} (\textbf{HS}) to understand how close our models perform in comparison to finetuned state-of-the-art models.\footnote{More details of \textbf{SC} and \textbf{HS} models can be found in \Sref{sec:finetune_details}}

From these Tables we see that in general as the size of the LM increases, the AUC and F1 performance for detecting bias improves.
We also see that \ours{} performs the best on both AUC and F1 metrics for all the SBIC tasks and it performs better than finetuned SC model on three tasks - \textit{offensive}, \textit{intent} and \textit{group}.
In Table~\ref{tab:sbic_results}, we see that using class balanced random sampling (\rndF) performs much better than random sampling.
We also observe that similarity based \tfidf{} sampling performs better than random (\rndF{}) sampling.
We note that as the model size increases, the improvement gained by using better sampling technique reduces. 
Concretely, across the four SBIC tasks, the average AUC gain between \rndF{} and \tfidf{} sampling is $9.6$\% for \mOne{}, $3.9$\% for \mEight{} and $3.4$\% for \ours{}.
This shows that the larger models are robust towards the sampling technique.

% From these Tables we see that in general as the size of the LM increases, the AUC and F1 performance for detecting bias improves.
%It also performs comparable to SC on the intent task.
For HOF task, \ours{} performs better than baselines on both F1 metrics.
For Target task, this is not the case because of the skewed distribution of indicative keywords in this task.
We perform an analysis of the percentage of posts that contain keywords and their correlation with labels (details in Appendix~\ref{sec:keyword_analysis}).
We note that keywords are present at five times higher rate in positive posts of the Target task compared to the other tasks.

The results for the two multi-class classification tasks are shown in Table~\ref{tab:multi_class_results}.
%\sm{Rafal: Please explain why we don't use KWD baseline for Table 3.}
For both the tasks we experiment with different values of $k$ ($k=\{7, 28\}$ for WHO task and $k=\{3, 12\}$ for HOP task) and we present the best performing results.
The results for WHO task are shown for $k=28$ shots, to ensure equal samples from each of the seven classes.
Similarly, the HOP results are shown for $k=3$ i.e one exemplar is picked from each of the three classes that is the closest to the textual post $\mathbf{Q}$.
For both the multi-class classifications tasks, \ours{} performs better than baselines on all metrics illustrating the effectiveness of our approach.
We are unable to show the keyword baselines for multi-class tasks because of lack of keyword repositories that clearly align with all the classes.

\section{Analysis and Discussion}
\label{sec:analysis}

\paragraph{Smaller Size of $\mathcal{D}$}
We experiment with smaller sizes of repository $\mathcal{D}$ to understand how the size of $\mathcal{D}$ affects model performance.
The goal is to understand the amount of annotated data required by our technique to detect social bias.
The original SBIC train set contains $\sim35$k instances (used as $\mathcal{D}$ in \Sref{sec:results}).
We down-sample examples from this dataset to create smaller sets of sizes $10$k, $1$k and $100$.
A labeled set $\mathcal{D}$ of size $100$ means that we can only select the $k$-shots from $100$ samples.
We ensure that we have equal label distribution in the down-sampled sets.
For each size, we use three different random seeds to generate down-sampled data.
We then average the results of the four $32$-shot SBIC classification tasks for the sets produced by three random seeds and present the results in Table~\ref{tab:train_set_results} for both \rndF{} and \tfidf{} sampling.

We observe that in case of \rndF{} sampling, there is practically no change in performance of the language models with the reduction in support repository size.
The standard deviation for the AUC score across the four $\mathcal{D}$ sizes is $0.43$ for \mOne{}, $0.54$ for \mEight{} and $0.47$ for \ours{}, which is extremely low.
We see that in case of \tfidf{} sampling, the performance of the larger models does not degrade substantially when the size of the labeled repository is reduced to as low as $100$ samples.
For example, the relative percentage AUC drop is only $2.93$\% for \mEight{} and $1.84$\% for \ours{} as opposed to $5.47$\% for the \mOne{} model.
The performance for the \tfidf{} baseline however drops substantially by AUC $10.6$\%.
For all the cases, \ours{} performs better than the baselines on both AUC and F1 metric.
For each support repository size, the LMs using \tfidf{} based sampling perform better than the corresponding LMs using \rndF{} sampling.

\paragraph{Contribution of Components}
We have four components in our input to model $\mathcal{M}$: the textual exemplars $\mathbf{x}_i$, the definition $\mathbf{d}$, the class label $\mathbf{c}_i$ of exemplar $\mathbf{x}_i$, and the textual query post $\mathbf{Q}$ which we want to classify.
We perform ablation studies with $\mathbf{x}_i$, $\mathbf{d}$ and $\mathbf{Q}$ to understand the contribution of each of them in making the final prediction.
Experiments with perturbation of label $\mathbf{c}_i$ are shown separately in the later section.
The results of $32$-shot intent classification ablation studies are shown in Table~\ref{tab:sbic_ablation}.

The AUC performance drops consistently as we remove the definition $\mathbf{d}$, the text of exemplars $\mathbf{x}_i$, and the query $\mathbf{Q}$.
For \mEight{} and \ours{}, we see that removing all three has the most impact on the performance.
The AUC drops on average by $19.67$\% across models by only removing the textual exemplars $\mathbf{x}_i$ (The model gets class labels $\mathbf{c}_i$ of exemplars, $\mathbf{d}$ and $\mathbf{Q}$ as input).
This suggests that \ours{} pays attention to all the components and the textual examples are as important as knowing their labels. 
Absence of definition has the smallest impact causing only a minor drop in AUC of $3.7$\% for \mOne{} and $2.43$\% for \ours{}, but no drop for \mEight{}.

%We get an improvement in F1 score for the \mOne{} model by removing the definition.

\begin{table}[t]
\centering
\small{
\begin{tabular}{@{}l c c c c r r @{}}
\textbf{Model} & \textbf{Post} & \textbf{Def.} & \textbf{Q} & \hspace{25.0pt}\textbf{AUC} & \textbf{F1} \\
\toprule
\mOne & \checkmark & \checkmark & \checkmark & \hspace{28.0pt} 63.20 & 69.04 \\
\mOne & \checkmark & \xmark & \checkmark & \hspace{4.0pt}\dar{3.70\%} 60.96 & 70.99 \\
\mOne & \xmark & \xmark &  \xmark &  \dar{19.00\%} 51.27 & 65.03 \\
\mOne & \xmark & \checkmark &  \checkmark & \dar{20.03\%} 50.62 & 61.27 \\
\mOne & \xmark & \checkmark &  \xmark & \dar{22.09\%} 49.32 & 52.32 \\
& \\
\mEight & \checkmark & \checkmark & \checkmark & \hspace{28.0pt} 66.49 & 74.39 \\
\mEight& \checkmark & \xmark & \checkmark & \hspace{4.0pt}\uab{1.76\%} 67.66 & 72.81  \\
\mEight & \xmark & \checkmark & \checkmark & \dar{20.80\%} 52.66 & 69.27  \\
\mEight & \xmark & \checkmark & \xmark & \dar{22.94\%} 51.24 & 67.05 \\
\mEight & \xmark & \xmark & \xmark & \dar{23.28\%} 51.01 & 66.69 \\
& \\
\ours &  \checkmark &  \checkmark & \checkmark & \hspace{28.0pt} 76.61 & 79.77 \\
\ours &  \checkmark & \xmark & \checkmark & \hspace{4.0pt}\dar{2.43\%} 74.75 & 73.37 \\
\ours &  \xmark & \checkmark & \checkmark & \dar{18.17\%} 52.69 & 70.44 \\
\ours &  \xmark & \checkmark & \xmark & \dar{34.64\%} 50.07 & 65.30 \\
\ours &  \xmark & \xmark & \xmark & \dar{34.51\%} 50.17 & 61.14 \\
\bottomrule
\end{tabular}
}
\vspace{-0.5em}
\caption{Ablation studies on SBIC \textit{Intent} task to understand the contribution of each component of the instruction. Ablations performed with $32$-shots prompting. The highest capacity model \text{\ours} achieves best performance with all the instruction components.}
%\vspace{-1.0em}
\label{tab:sbic_ablation}
\end{table}

\paragraph{Robustness to Labels}
To understand the contribution of labels in the $k$-shot binary classification performance, we perform two ablation studies.
We flip the labels (\textbf{Flip}) of the exemplars i.e. if the ground truth label of exemplar $\mathbf{x}_i$ is $c_i=$\textit{``Yes"}, then we supply the label \textit{``No"} and vice versa.
This study is done to understand the robustness of the model to noisy or wrong labels.
In the second experiment, we flip the labels only $50$\% (\textbf{Random}) of the times.
Hence, $50$\% of the time model gets correct label to an exemplar and gets a wrong label the other $50$\% of time.
We show the results on $32$-shot intent classification task for both \tfidf{} and \rndF{} sampling in Table~\ref{tab:flipped_lab_results}.

We observe that the AUC and F1 accuracy drops the most when we supply flipped labels.
Interestingly, even in this case, the language models identify intentional insult more than a third of the time.
% and their performance does not go down to zero
% \rk{Interestingly, even in this case, the \ours{} model identifies intentional insult more than a third of the time and with much higher AUC of 38.10, than expected from an inverse performance compared to the original task (AUC of 23.39). This means that the model has some ``resistance'' towards learning labels in the opposite direction.}
In case of supplying Random labels, the relative AUC performance of LMs drops on average by $16$\% (similar results shown for general NLP tasks \cite{song2022learning}) further showcasing the robustness of the models towards wrong labels.
The \rndF{} sampling is in general more robust to label flips compared to the \tfidf{} sampling.
For example, on average for the Random experiment, the AUC performance across LMs drops by $13.6$\% for \rndF{} sampling and $18.6$\% for \tfidf{} sampling.
\tfidf{} picks the shots carefully to augment the LMs ability to classify and hence when wrong labels are provided, this sampling technique suffers more loss compared to \rndF{}.

% The performance does not drop significantly in the Random study further showcasing the robustness of the models towards wrong labels.
% This study shows that as long as the model is supplied with some exemplars with correct labels, it performs well.

\begin{table}[t]
\centering
\small{
\begin{tabular}{@{}l l r r r@{}}
\textbf{Model} & \textbf{Sam} & \multicolumn{1}{c}{\textbf{AUC}} & \multicolumn{1}{c}{\textbf{F1}} & \textbf{Experiment} \\
\toprule
\mOne & \tfidf & 42.75 & 49.63 & Flip \\
\mOne & \tfidf & 52.61 & 59.23 & Random \\
\mOne & \tfidf & 63.20 & 69.04 & Correct \\
& \\
\mEight & \tfidf & 39.26 & 51.37 & Flip \\
\mEight & \tfidf & 54.02 & 63.88 & Random \\
\mEight & \tfidf & 66.49 & 74.39 & Correct \\
& \\
\ours & \tfidf & 38.10 & 41.80 & Flip \\
\ours & \tfidf & 61.08 & 65.18 & Random \\
\ours & \tfidf & 76.61 & 79.77 & Correct \\
\midrule
\mOne & \rndF & 44.46 & 36.07 & Flip \\
\mOne & \rndF & 50.81 & 47.97 & Random \\
\mOne & \rndF & 58.04 & 57.79 & Correct \\
& \\
\mEight & \rndF & 48.45 & 57.18 & Flip \\
\mEight & \rndF & 56.57 & 64.31 & Random \\
\mEight & \rndF & 64.88 & 73.05 & Correct \\
& \\
\ours & \rndF & 44.68 & 43.45 & Flip \\
\ours & \rndF & 63.79 & 66.20 & Random \\
\ours & \rndF & 75.63 & 78.38 & Correct \\
\bottomrule
\end{tabular}
}
\vspace{-0.5em}
\caption{Experiment to understand the role of labels in prediction with $32$-shot prompts on SBIC \textit{Intent} task. The \textit{Flip} and \textit{Random} denote reversing the labels or replacing half of the shots with random label respectively.} 
\vspace{-1.0em}
\label{tab:flipped_lab_results}
\end{table}

\paragraph{$k$-shots vs Metric}

\begin{figure}[!t]
\centering
{
\includegraphics[width=\linewidth]{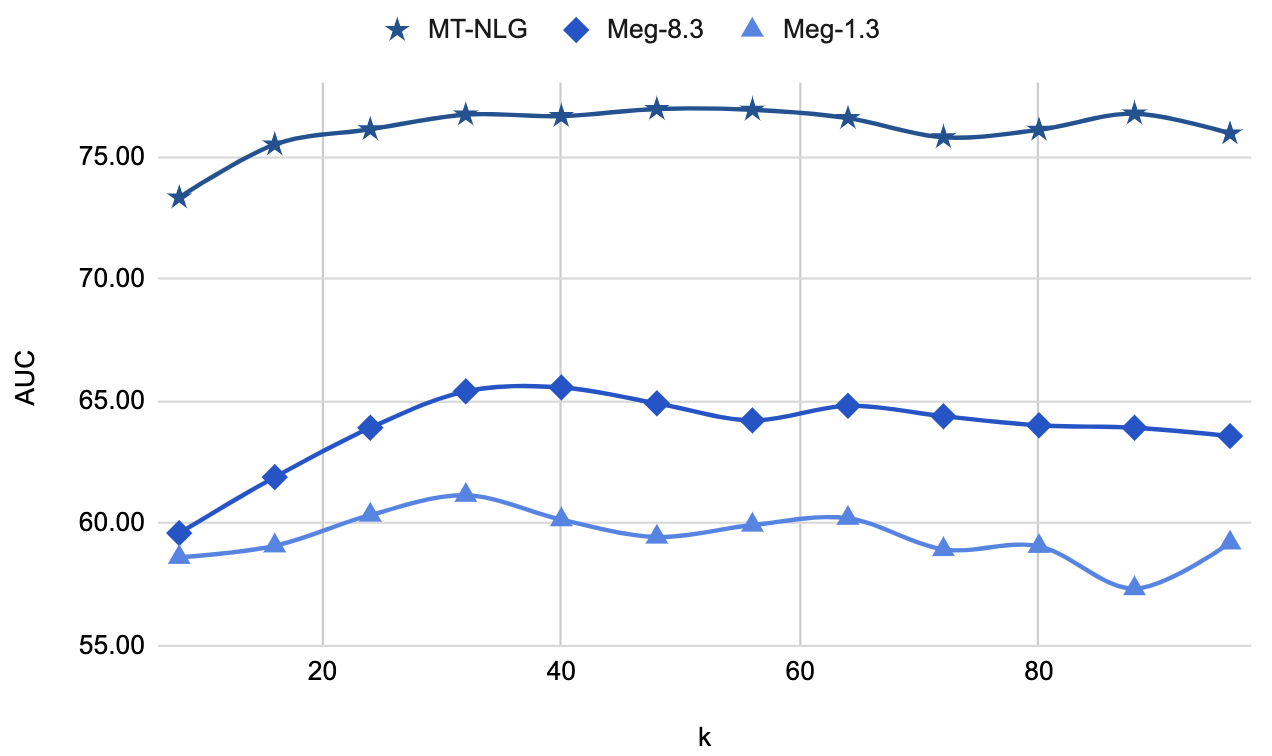}
}
\vspace{-1.5em}
\caption{Experiment with varying number of instruction shots for the \textit{Group} binary classification task from SBIC. Change in \textit{AUC} scores is plotted against the number of shots $k$ provided as context to the LMs.}
\vspace{-1.0em}
\label{fig:k_shots}
\end{figure}

To understand how the performance of the models scale with number of shots, we present Figure~\ref{fig:k_shots}.
It shows the graph of AUC metric for the group classification task with number of shots ranging from $8$ to $96$ with a step size of $8$.
We observe that for the group task, the performance improves up to $k$  in range of $32-48$ and then plateaus.
The best performance of \ours{} is at $k=48$ with AUC score of $76.96$ which is a $0.23$ AUC improvement over $k=32$.
We would like to note that the $k$ at which the model achieves optimal performance is task dependent.
For uniformity we choose to report performance at $k=32$ for all tasks but we believe that each category can be further improved.

\paragraph{Qualitative Analysis}

%To better understand the predictions of \ours{}, 
We randomly sample $50$ cases where \ours{} makes a wrong prediction and \tfidf{} is correct.
Similarly, we also sample $50$ cases where \ours{} predicts the correct label and \tfidf{} makes correct predictions.
All the samples are picked for the offensive classification task.
Based on manual qualitative coding among two of the authors, we further divide these cases into five categories.\footnote{Example posts for each category are shown in Table~\ref{tab:qualitative_analysis} in Appendix (Warning: the examples shown contain content that maybe offensive or upsetting).}
%which are presented in Figure~\ref{fig:disagreement_classes}.

Overall, we observe that for both \tfidf{} and \ours{}, high number of errors occur when there are no keywords\footnote{based on \emph{LDNOOBW} dataset \cite{GitHubLD98:online}} present in the query text.
In case of offensive posts without keywords, \ours{} makes $12$\% less errors compared to \tfidf{}.
This shows that \tfidf{} finds it challenging to identify implicit bias in the text.
We observe that some posts may contain trigger words such as \textit{white nationalist}, \textit{Black}, \textit{Jews}, \textit{Muslims}, \textit{9/11}, etc. which causes \tfidf{} to retrieve shots with offensive keywords from $\mathcal{D}$.
We also observe that \tfidf{} picks irrelevant offensive shots when enough content on the same topic is not available in $\mathcal{D}$.
When such irrelevant posts are used, \tfidf{} makes a wrong prediction.
When keywords are present in the posts, there are lesser number of errors.
Some of the posts are not offensive but contain slurs such as \textit{f**k}, \textit{b***h}, etc.

Finally, $5$\% instances are mislabeled according to the human annotators.\footnote{Note that we do not claim that $5$\% of the entire corpus is mislabeled.}
Some posts were clearly offensive and were annotated as non-offensive whereas other posts that seem to be non-offensive were marked as offensive suggesting potential ambiguity of interpretation.
In these cases, there was not enough context to make a decision, a limitation particularly relevant to social media datasets and identified in prior work \cite{chen2018using}.

\begin{figure}[!t]
\centering
{
\includegraphics[width=\linewidth]{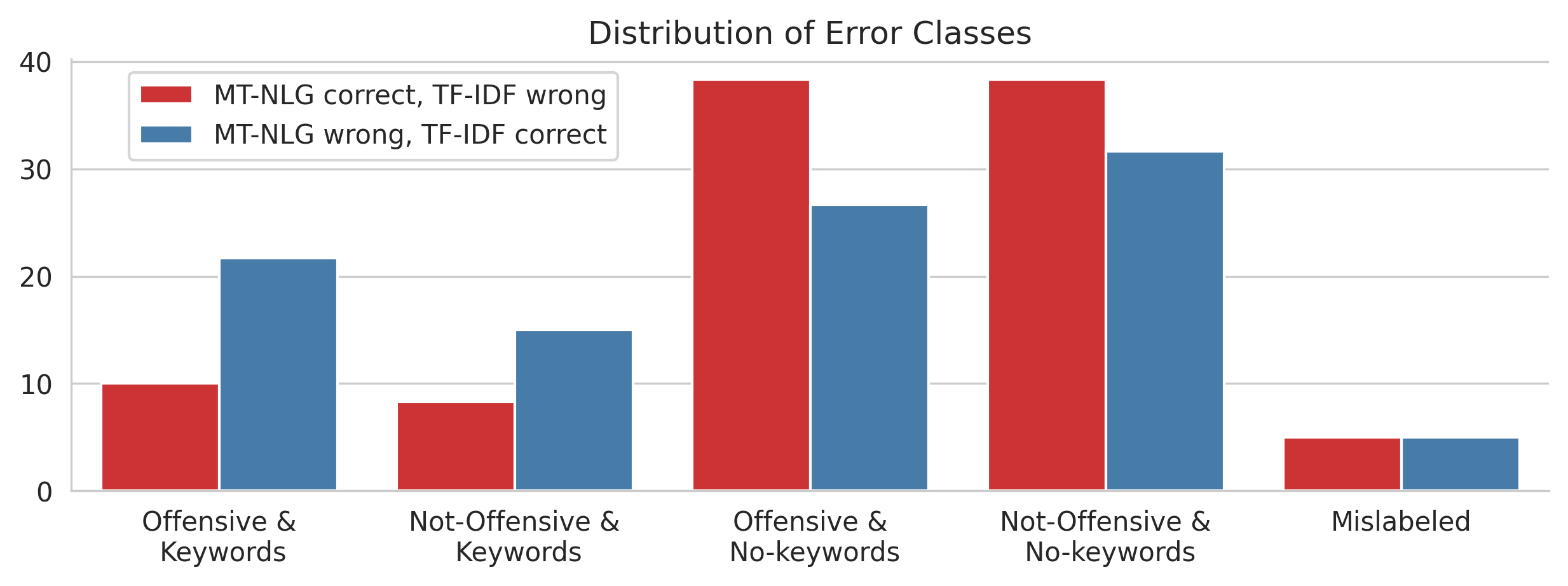}
}
\vspace{-1.5em}
\caption{Error classes from qualitative analysis of disagreements between \tfidf{} and \ours{} on \textit{Offensive} task from SBIC. The higher percentage of correct classifications even in cases of missing explicit keywords by \ours{} compared to \tfidf{} (red bars) demonstrates that \ours{} better captures implicit bias.}
% Old: Fewer misclassifications due to lack of explicit keywords by \ours{} compared to \tfidf{} (red bars) demonstrates that \ours{} better captures implicit bias.}
\vspace{-1.0em}
\label{fig:disagreement_classes}
\end{figure}

\section{Related Work}
\label{sec:related_work}

\paragraph{Bias Detection}
Approaches for detecting bias in text can broadly be put into 3 categories: 1) keyword-based, 2) feature-representation-based, 3) supervised-training-based. 

Keywords-based methods rely on curated lexicons of words  \cite{Hatebase11:online,GitHubLD98:online,UsefulRe32:online}.
%These have been extensively used in prior research \cite{sood2012automatic, mondal2017measurement} as well as in commercial settings \cite{AIandthe49:online}.
Despite wide use \cite{sood2012automatic, mondal2017measurement,AIandthe49:online} they can be at a significant mismatch with human ratings \cite{hateoffensive} and can't easily capture phenomenons such as sarcasm, humor \cite{rajadesingan2015sarcasm} or polysemy
%(i.e., different meaning in different context, e.g., ``white trash'' vs. ``white trash cans'') 
\cite{sahlgren2018learning}.
Such lexicons require constant updates as new slang develops \cite{nobata2016abusive}. 

% Feature-representation methods
Prior work has also explored sophisticated feature representations including n-grams, linguistic and syntactic features \cite{nobata2016abusive}, \emph{TF-IDF} \cite{salminen2018anatomy}, Bag of Words and word embeddings \cite{djuric2015hate}, as well as content-specific features such as mentions, proper nouns, named entities, and target group specific vocabularies \cite{waseem2017understanding}.
Topic modeling approaches such as \emph{Labeled Latent Dirichlet Allocation} have also been proposed \cite{saleem2017web}. 
Overall, these methods try to provide better feature representations than keywords \cite{sahlgren2018learning}, but rely on careful feature engineering which might be specific to particular context and hence makes them inflexible across settings. 

% Training-based methods
Supervised training based methods rely on large labeled datasets to train models \cite{badjatiya2017deep,pavlopoulos2017deeper,zhang2018detecting,d2019towards,caselli2020hatebert,silva2020data}.
%from classical machine learning \cite{badjatiya2017deep}, deep learning such as RNNs \cite{pavlopoulos2017deeper}, CNNs \cite{zhang2018detecting} or BERT and GPT-2 transformers \cite{d2019towards, caselli2020hatebert, silva2020data}. 
Recently, transformers have been fine-tuned for hate-speech detection \cite{caselli2020hatebert}, detection of targeted offensive language \cite{rosenthal2021solid} and bias \cite{sap-etal-2020-social}.
Similar techniques with focus on toxicity have been adopted in the commercial \emph{Perspective API} \cite{Perspect22:online}.
These methods are more effective than keywords or custom features \cite{badjatiya2017deep}, but they rely on large labeled datasets and are expensive, or even impossible (e.g., Perspective API) to retrain. 
Furthermore, most approaches focus on binary coarse-grained hate-speech or toxicity classification, rather than on nuanced and target group specific issues of bias.

%\cite{hateoffensive} automatically generated a lexicon of offensive words based on labeled dataset. \cite{van2021grievance} A Grievance Dictionary developed in \cite{van2021grievance} can be used for grievance-fueled violence threat assessment. Similarly Linguistic Inquiry and Word Count (LIWC) is one of the most prominent psycholinguistic dictionaries developed for  (LIWC; Pennebaker et al., 2015).

\paragraph{Prompting}
Recent success of large pre-trained language models~\cite{devlin-etal-2019-bert,brown-etal-2020-gpt3,Patwary2021using} has opened the field to the new direction of prompting~\cite{liu2021pre} them for various NLP tasks.
We demonstrate the success of this technique on social bias detection tasks which makes it possible to detect different social biases without a huge labeled set and training separate models for each task.
\citet{schick2021self} self-diagnose toxicity in machine generated text.
This is closest to our work.
We focus on bias and a more fine-grained understanding of bias such as the target group, intentional or unintentional offense etc. in human written text.
Bias in human written text can be more challenging to detect as it can be riddled with sarcasm and humor.
\citet{schick2021self} evaluate the success of large LMs to detect toxicity using the scores provided by automated classifier Perspective API~\cite{Perspect22:online} as the ground truth.
We evaluate the success of large LMs at detecting bias using human annotated labels as ground truth.
Most importantly, \citet{schick2021self} sample the most toxic and most non-toxic examples from RealToxicity~\cite{gehman2020realtoxicityprompts} dataset i.e. they sample the extreme cases of toxicity, and report the performance of large LMs on zero-shot classification.
We report few-shot classification performance on the entire test set of two independent datasets with human-provided labeling.
Additionally, we also showcase multi-class classification capability of our approach.

\citet{liu2021makes} investigate the retrieval of exemplars that are semantically-similar to a test sample.
The two key differences of our work are: (1) we focus on few-shot instruction based bias detection, and (2) we select equal number of exemplars from each class.
%\citet{liu2021makes} choose pre-trained language models such as BERT~\cite{devlin-etal-2019-bert} and RoBERTa~\cite{liu2019roberta} as sentence encoders, whereas we use Sentence-BERT~\cite{reimers-2019-sentence-bert} specializing in retrieving semantically similar sentences.

\section{Conclusion}
\label{sec:conclusion}
The paper proposes a novel technique to select exemplars for few-shot instruction-based method to detect bias using pretrained model's internal knowledge and no fine-tuning. 
On two separate datasets involving binary and multi-class classification (ranging from coarse-grained to fine-grained tasks), we demonstrate that sufficiently large pretrained LMs can detect bias beating keyword and semantic-based heuristics as well as fine-tuned models on some of the tasks. 
In subsequent experiments we show that our method is: 
1) flexible in incorporating various bias definitions, 
2) robust against small pool of labeled documents to select shots from, 
3) relies on deeper semantic interpretation, rather than surface level heuristics, which we show through extensive ablation studies and manual qualitative inspection.
The ethical considerations are discussed in Appendix \ref{sec:ethics}.

%and even with best intentions there will be disagreement among humans around the interpretations \cite{chen2018using}.

%Such mislabeling can be intentional (to confuse the detection) or unintentional (e.g., unrealized bias of the annotators).

% Misuse - intentional or unintentional
% - undesired uses - censorship and banishing expression
% - purposeful selection of misleading labels in shots
% - not banning legitimate use of violent language - e.g., (crime fiction)
%

% - performance metrics optimization
% - fairness towards different demographics
% - 

% limitations
% - ambiguity in interpretation due to lack of context

%\begin{itemize}
%    \item intended use
%    \item misuse
%    \item how labeling is noisy and how that affects the %prediction
%\end{itemize}

% \section*{Acknowledgements}
% This material includes work supported by the National Science Foundation under Grant \#2030859 to the Computing Research Association for the CIFellows Project. 
% Any opinions, findings, and conclusions or recommendations expressed in this material are those of the author(s) and do not necessarily reflect the views of the National Science Foundation nor the Computing Research Association.

\bibliography{anthology,custom}

\begin{thebibliography}{66}
\expandafter\ifx\csname natexlab\endcsname\relax\def\natexlab#1{#1}\fi

\bibitem[{Badjatiya et~al.(2017)Badjatiya, Gupta, Gupta, and
  Varma}]{badjatiya2017deep}
Pinkesh Badjatiya, Shashank Gupta, Manish Gupta, and Vasudeva Varma. 2017.
\newblock Deep learning for hate speech detection in tweets.
\newblock In \emph{Proceedings of the 26th international conference on World
  Wide Web companion}, pages 759--760.

\bibitem[{Brown et~al.(2020)Brown, Mann, Ryder, Subbiah, Kaplan, Dhariwal,
  Neelakantan, Shyam, Sastry, Askell, Agarwal, Herbert{-}Voss, Krueger,
  Henighan, Child, Ramesh, Ziegler, Wu, Winter, Hesse, Chen, Sigler, Litwin,
  Gray, Chess, Clark, Berner, McCandlish, Radford, Sutskever, and
  Amodei}]{brown-etal-2020-gpt3}
Tom~B. Brown, Benjamin Mann, Nick Ryder, Melanie Subbiah, Jared Kaplan,
  Prafulla Dhariwal, Arvind Neelakantan, Pranav Shyam, Girish Sastry, Amanda
  Askell, Sandhini Agarwal, Ariel Herbert{-}Voss, Gretchen Krueger, Tom
  Henighan, Rewon Child, Aditya Ramesh, Daniel~M. Ziegler, Jeffrey Wu, Clemens
  Winter, Christopher Hesse, Mark Chen, Eric Sigler, Mateusz Litwin, Scott
  Gray, Benjamin Chess, Jack Clark, Christopher Berner, Sam McCandlish, Alec
  Radford, Ilya Sutskever, and Dario Amodei. 2020.
\newblock \href {http://arxiv.org/abs/2005.14165} {Language models are few-shot
  learners}.
\newblock \emph{CoRR}, abs/2005.14165.

\bibitem[{Caselli et~al.(2020)Caselli, Basile, Mitrovi{\'c}, and
  Granitzer}]{caselli2020hatebert}
Tommaso Caselli, Valerio Basile, Jelena Mitrovi{\'c}, and Michael Granitzer.
  2020.
\newblock Hatebert: Retraining bert for abusive language detection in english.
\newblock \emph{arXiv preprint arXiv:2010.12472}.

\bibitem[{Chen et~al.(2018)Chen, Drouhard, Kocielnik, Suh, and
  Aragon}]{chen2018using}
Nan-Chen Chen, Margaret Drouhard, Rafal Kocielnik, Jina Suh, and Cecilia~R
  Aragon. 2018.
\newblock Using machine learning to support qualitative coding in social
  science: Shifting the focus to ambiguity.
\newblock \emph{ACM Transactions on Interactive Intelligent Systems (TiiS)},
  8(2):1--20.

\bibitem[{Chung et~al.(2019)Chung, Kuzmenko, Tekiroglu, and
  Guerini}]{chung-etal-2019-conan}
Yi-Ling Chung, Elizaveta Kuzmenko, Serra~Sinem Tekiroglu, and Marco Guerini.
  2019.
\newblock \href {https://doi.org/10.18653/v1/P19-1271} {{CONAN} - {CO}unter
  {NA}rratives through nichesourcing: a multilingual dataset of responses to
  fight online hate speech}.
\newblock In \emph{Proceedings of the 57th Annual Meeting of the Association
  for Computational Linguistics}, pages 2819--2829, Florence, Italy.
  Association for Computational Linguistics.

\bibitem[{Davidson et~al.(2019)Davidson, Bhattacharya, and
  Weber}]{davidson2019racial}
Thomas Davidson, Debasmita Bhattacharya, and Ingmar Weber. 2019.
\newblock Racial bias in hate speech and abusive language detection datasets.
\newblock \emph{arXiv preprint arXiv:1905.12516}.

\bibitem[{Davidson et~al.(2017{\natexlab{a}})Davidson, Warmsley, Macy, and
  Weber}]{davidson2017automated}
Thomas Davidson, Dana Warmsley, Michael Macy, and Ingmar Weber.
  2017{\natexlab{a}}.
\newblock Automated hate speech detection and the problem of offensive
  language.
\newblock In \emph{Proceedings of the International AAAI Conference on Web and
  Social Media}, volume~11.

\bibitem[{Davidson et~al.(2017{\natexlab{b}})Davidson, Warmsley, Macy, and
  Weber}]{hateoffensive}
Thomas Davidson, Dana Warmsley, Michael Macy, and Ingmar Weber.
  2017{\natexlab{b}}.
\newblock Automated hate speech detection and the problem of offensive
  language.
\newblock In \emph{Proceedings of the 11th International AAAI Conference on Web
  and Social Media}, ICWSM '17, pages 512--515.

\bibitem[{Davison et~al.(2019)Davison, Feldman, and
  Rush}]{davison-etal-2019-commonsense}
Joe Davison, Joshua Feldman, and Alexander~M Rush. 2019.
\newblock Commonsense knowledge mining from pretrained models.
\newblock In \emph{Proceedings of the 2019 Conference on Empirical Methods in
  Natural Language Processing and the 9th International Joint Conference on
  Natural Language Processing (EMNLP-IJCNLP)}, pages 1173--1178.

\bibitem[{Devlin et~al.(2019)Devlin, Chang, Lee, and
  Toutanova}]{devlin-etal-2019-bert}
Jacob Devlin, Ming-Wei Chang, Kenton Lee, and Kristina Toutanova. 2019.
\newblock \href {https://doi.org/10.18653/v1/N19-1423} {{BERT}: Pre-training of
  deep bidirectional transformers for language understanding}.
\newblock In \emph{Proceedings of the 2019 Conference of the North {A}merican
  Chapter of the Association for Computational Linguistics: Human Language
  Technologies, Volume 1 (Long and Short Papers)}, pages 4171--4186,
  Minneapolis, Minnesota. Association for Computational Linguistics.

\bibitem[{Djuric et~al.(2015)Djuric, Zhou, Morris, Grbovic, Radosavljevic, and
  Bhamidipati}]{djuric2015hate}
Nemanja Djuric, Jing Zhou, Robin Morris, Mihajlo Grbovic, Vladan Radosavljevic,
  and Narayan Bhamidipati. 2015.
\newblock Hate speech detection with comment embeddings.
\newblock In \emph{Proceedings of the 24th international conference on world
  wide web}, pages 29--30.

\bibitem[{Does et~al.(2011)Does, Derks, and Ellemers}]{does2011thou}
Serena Does, Belle Derks, and Naomi Ellemers. 2011.
\newblock Thou shalt not discriminate: How emphasizing moral ideals rather than
  obligations increases whites' support for social equality.
\newblock \emph{Journal of Experimental Social Psychology}, 47(3):562--571.

\bibitem[{D'sa et~al.(2019)D'sa, Illina, and Fohr}]{d2019towards}
Ashwin~Geet D'sa, Irina Illina, and Dominique Fohr. 2019.
\newblock Towards non-toxic landscapes: Automatic toxic comment detection using
  dnn.
\newblock \emph{arXiv preprint arXiv:1911.08395}.

\bibitem[{Fatemi et~al.(2021)Fatemi, Xing, Liu, and
  Xiong}]{fatemi2021improving}
Zahra Fatemi, Chen Xing, Wenhao Liu, and Caiming Xiong. 2021.
\newblock Improving gender fairness of pre-trained language models without
  catastrophic forgetting.
\newblock \emph{arXiv preprint arXiv:2110.05367}.

\bibitem[{Fiske(1993)}]{fiske1993controlling}
Susan~T Fiske. 1993.
\newblock Controlling other people: The impact of power on stereotyping.
\newblock \emph{American psychologist}, 48(6):621.

\bibitem[{Flekova et~al.(2016)Flekova, Carpenter, Giorgi, Ungar, and
  Preo{\c{t}}iuc-Pietro}]{flekova2016analyzing}
Lucie Flekova, Jordan Carpenter, Salvatore Giorgi, Lyle Ungar, and Daniel
  Preo{\c{t}}iuc-Pietro. 2016.
\newblock Analyzing biases in human perception of user age and gender from
  text.
\newblock In \emph{Proceedings of the 54th Annual Meeting of the Association
  for Computational Linguistics (Volume 1: Long Papers)}, pages 843--854.

\bibitem[{Gao et~al.(2020)Gao, Fisch, and Chen}]{gao2020making}
Tianyu Gao, Adam Fisch, and Danqi Chen. 2020.
\newblock Making pre-trained language models better few-shot learners.
\newblock \emph{arXiv preprint arXiv:2012.15723}.

\bibitem[{Gehman et~al.(2020)Gehman, Gururangan, Sap, Choi, and
  Smith}]{gehman2020realtoxicityprompts}
Samuel Gehman, Suchin Gururangan, Maarten Sap, Yejin Choi, and Noah~A Smith.
  2020.
\newblock Realtoxicityprompts: Evaluating neural toxic degeneration in language
  models.
\newblock In \emph{Findings of the Association for Computational Linguistics:
  EMNLP 2020}, pages 3356--3369.

\bibitem[{Hashimoto et~al.(2017)Hashimoto, Xiong, Tsuruoka, and
  Socher}]{hashimoto2017joint}
Kazuma Hashimoto, Caiming Xiong, Yoshimasa Tsuruoka, and Richard Socher. 2017.
\newblock A joint many-task model: Growing a neural network for multiple nlp
  tasks.
\newblock In \emph{Proceedings of the 2017 Conference on Empirical Methods in
  Natural Language Processing}, pages 1923--1933.

\bibitem[{Hatebase.org(2021)}]{Hatebase11:online}
Hatebase.org. 2021.
\newblock Hatebase.
\newblock \url{https://hatebase.org/}.
\newblock (Accessed on 12/08/2021).

\bibitem[{Hovy and Prabhumoye(2021)}]{hovy2021five}
Dirk Hovy and Shrimai Prabhumoye. 2021.
\newblock Five sources of bias in natural language processing.
\newblock \emph{Language and Linguistics Compass}, 15(8):e12432.

\bibitem[{Liu et~al.(2021{\natexlab{a}})Liu, Shen, Zhang, Dolan, Carin, and
  Chen}]{liu2021makes}
Jiachang Liu, Dinghan Shen, Yizhe Zhang, Bill Dolan, Lawrence Carin, and Weizhu
  Chen. 2021{\natexlab{a}}.
\newblock What makes good in-context examples for gpt-$3 $?
\newblock \emph{arXiv preprint arXiv:2101.06804}.

\bibitem[{Liu et~al.(2021{\natexlab{b}})Liu, Yuan, Fu, Jiang, Hayashi, and
  Neubig}]{liu2021pre}
Pengfei Liu, Weizhe Yuan, Jinlan Fu, Zhengbao Jiang, Hiroaki Hayashi, and
  Graham Neubig. 2021{\natexlab{b}}.
\newblock Pre-train, prompt, and predict: A systematic survey of prompting
  methods in natural language processing.
\newblock \emph{arXiv preprint arXiv:2107.13586}.

\bibitem[{Logan~IV et~al.(2021)Logan~IV, Bala{\v{z}}evi{\'c}, Wallace, Petroni,
  Singh, and Riedel}]{logan2021cutting}
Robert~L Logan~IV, Ivana Bala{\v{z}}evi{\'c}, Eric Wallace, Fabio Petroni,
  Sameer Singh, and Sebastian Riedel. 2021.
\newblock Cutting down on prompts and parameters: Simple few-shot learning with
  language models.
\newblock \emph{arXiv preprint arXiv:2106.13353}.

\bibitem[{Lohr(2021)}]{lohr2021sampling}
Sharon~L Lohr. 2021.
\newblock \emph{Sampling: design and analysis}.
\newblock Chapman and Hall/CRC.

\bibitem[{Mandl et~al.(2019)Mandl, Modha, Majumder, Patel, Dave, Mandlia, and
  Patel}]{mandl-etal-2019-hasoc}
Thomas Mandl, Sandip Modha, Prasenjit Majumder, Daksh Patel, Mohana Dave,
  Chintak Mandlia, and Aditya Patel. 2019.
\newblock \href {https://doi.org/10.1145/3368567.3368584} {Overview of the
  hasoc track at fire 2019: Hate speech and offensive content identification in
  indo-european languages}.
\newblock In \emph{Proceedings of the 11th Forum for Information Retrieval
  Evaluation}, FIRE '19, page 14–17, New York, NY, USA. Association for
  Computing Machinery.

\bibitem[{Mondal et~al.(2017)Mondal, Silva, and
  Benevenuto}]{mondal2017measurement}
Mainack Mondal, Leandro~Ara{\'u}jo Silva, and Fabr{\'\i}cio Benevenuto. 2017.
\newblock A measurement study of hate speech in social media.
\newblock In \emph{Proceedings of the 28th ACM conference on hypertext and
  social media}, pages 85--94.

\bibitem[{Nie et~al.(2020)Nie, Williams, Dinan, Bansal, Weston, and
  Kiela}]{nie2020adversarial}
Yixin Nie, Adina Williams, Emily Dinan, Mohit Bansal, Jason Weston, and Douwe
  Kiela. 2020.
\newblock Adversarial nli: A new benchmark for natural language understanding.
\newblock In \emph{Proceedings of the 58th Annual Meeting of the Association
  for Computational Linguistics}, pages 4885--4901.

\bibitem[{Nobata et~al.(2016)Nobata, Tetreault, Thomas, Mehdad, and
  Chang}]{nobata2016abusive}
Chikashi Nobata, Joel Tetreault, Achint Thomas, Yashar Mehdad, and Yi~Chang.
  2016.
\newblock Abusive language detection in online user content.
\newblock In \emph{Proceedings of the 25th international conference on world
  wide web}, pages 145--153.

\bibitem[{Pavlopoulos et~al.(2017)Pavlopoulos, Malakasiotis, and
  Androutsopoulos}]{pavlopoulos2017deeper}
John Pavlopoulos, Prodromos Malakasiotis, and Ion Androutsopoulos. 2017.
\newblock Deeper attention to abusive user content moderation.
\newblock In \emph{Proceedings of the 2017 conference on empirical methods in
  natural language processing}, pages 1125--1135.

\bibitem[{PerspectiveAPI(2021)}]{Perspect22:online}
PerspectiveAPI. 2021.
\newblock Perspective | developers.
\newblock \url{https://developers.perspectiveapi.com/s/}.
\newblock (Accessed on 12/08/2021).

\bibitem[{Petroni et~al.(2019)Petroni, Rockt{\"a}schel, Riedel, Lewis, Bakhtin,
  Wu, and Miller}]{petroni-etal-2019-language}
Fabio Petroni, Tim Rockt{\"a}schel, Sebastian Riedel, Patrick Lewis, Anton
  Bakhtin, Yuxiang Wu, and Alexander Miller. 2019.
\newblock \href {https://doi.org/10.18653/v1/D19-1250} {Language models as
  knowledge bases?}
\newblock In \emph{Proceedings of the 2019 Conference on Empirical Methods in
  Natural Language Processing and the 9th International Joint Conference on
  Natural Language Processing (EMNLP-IJCNLP)}, pages 2463--2473, Hong Kong,
  China. Association for Computational Linguistics.

\bibitem[{Rajadesingan et~al.(2015)Rajadesingan, Zafarani, and
  Liu}]{rajadesingan2015sarcasm}
Ashwin Rajadesingan, Reza Zafarani, and Huan Liu. 2015.
\newblock Sarcasm detection on twitter: A behavioral modeling approach.
\newblock In \emph{Proceedings of the eighth ACM international conference on
  web search and data mining}, pages 97--106.

\bibitem[{Reimers and Gurevych(2019)}]{reimers-2019-sentence-bert}
Nils Reimers and Iryna Gurevych. 2019.
\newblock \href {https://arxiv.org/abs/1908.10084} {Sentence-bert: Sentence
  embeddings using siamese bert-networks}.
\newblock In \emph{Proceedings of the 2019 Conference on Empirical Methods in
  Natural Language Processing}. Association for Computational Linguistics.

\bibitem[{Rosenthal et~al.(2021)Rosenthal, Atanasova, Karadzhov, Zampieri, and
  Nakov}]{rosenthal2021solid}
Sara Rosenthal, Pepa Atanasova, Georgi Karadzhov, Marcos Zampieri, and Preslav
  Nakov. 2021.
\newblock Solid: A large-scale semi-supervised dataset for offensive language
  identification.
\newblock In \emph{Findings of the Association for Computational Linguistics:
  ACL-IJCNLP 2021}, pages 915--928.

\bibitem[{Ruder(2017)}]{ruder2017overview}
Sebastian Ruder. 2017.
\newblock An overview of multi-task learning in deep neural networks.
\newblock \emph{arXiv preprint arXiv:1706.05098}.

\bibitem[{Sahlgren et~al.(2018)Sahlgren, Isbister, and
  Olsson}]{sahlgren2018learning}
Magnus Sahlgren, Tim Isbister, and Fredrik Olsson. 2018.
\newblock Learning representations for detecting abusive language.
\newblock In \emph{Proceedings of the 2nd Workshop on Abusive Language Online
  (ALW2)}, pages 115--123.

\bibitem[{Saleem et~al.(2017)Saleem, Dillon, Benesch, and
  Ruths}]{saleem2017web}
Haji~Mohammad Saleem, Kelly~P Dillon, Susan Benesch, and Derek Ruths. 2017.
\newblock A web of hate: Tackling hateful speech in online social spaces.
\newblock \emph{arXiv preprint arXiv:1709.10159}.

\bibitem[{Salminen et~al.(2019)Salminen, Almerekhi, Kamel, Jung, and
  Jansen}]{salminen2019online}
Joni Salminen, Hind Almerekhi, Ahmed~Mohamed Kamel, Soon-gyo Jung, and
  Bernard~J Jansen. 2019.
\newblock Online hate ratings vary by extremes: A statistical analysis.
\newblock In \emph{Proceedings of the 2019 Conference on Human Information
  Interaction and Retrieval}, pages 213--217.

\bibitem[{Salminen et~al.(2018)Salminen, Almerekhi, Milenkovi{\'c}, Jung, An,
  Kwak, and Jansen}]{salminen2018anatomy}
Joni Salminen, Hind Almerekhi, Milica Milenkovi{\'c}, Soon-gyo Jung, Jisun An,
  Haewoon Kwak, and Bernard~J Jansen. 2018.
\newblock Anatomy of online hate: developing a taxonomy and machine learning
  models for identifying and classifying hate in online news media.
\newblock In \emph{Twelfth International AAAI Conference on Web and Social
  Media}.

\bibitem[{Sap et~al.(2019)Sap, Card, Gabriel, Choi, and Smith}]{sap2019risk}
Maarten Sap, Dallas Card, Saadia Gabriel, Yejin Choi, and Noah~A Smith. 2019.
\newblock The risk of racial bias in hate speech detection.
\newblock In \emph{Proceedings of the 57th annual meeting of the association
  for computational linguistics}, pages 1668--1678.

\bibitem[{Sap et~al.(2020)Sap, Gabriel, Qin, Jurafsky, Smith, and
  Choi}]{sap-etal-2020-social}
Maarten Sap, Saadia Gabriel, Lianhui Qin, Dan Jurafsky, Noah~A. Smith, and
  Yejin Choi. 2020.
\newblock \href {https://doi.org/10.18653/v1/2020.acl-main.486} {Social bias
  frames: Reasoning about social and power implications of language}.
\newblock In \emph{Proceedings of the 58th Annual Meeting of the Association
  for Computational Linguistics}, pages 5477--5490, Online. Association for
  Computational Linguistics.

\bibitem[{Schick et~al.(2021)Schick, Udupa, and Sch{\"u}tze}]{schick2021self}
Timo Schick, Sahana Udupa, and Hinrich Sch{\"u}tze. 2021.
\newblock Self-diagnosis and self-debiasing: A proposal for reducing
  corpus-based bias in nlp.
\newblock \emph{arXiv preprint arXiv:2103.00453}.

\bibitem[{Scikit-learn(2022{\natexlab{a}})}]{AUCSciKit:online}
Scikit-learn. 2022{\natexlab{a}}.
\newblock Roc-auc-score.
\newblock
  \url{https://scikit-learn.org/stable/modules/generated/sklearn.metrics.roc_auc_score.html}.
\newblock (Accessed on 04/13/2022).

\bibitem[{Scikit-learn(2022{\natexlab{b}})}]{TfIdfSciKit:online}
Scikit-learn. 2022{\natexlab{b}}.
\newblock Tfidfvectorizer.
\newblock \url{https://tinyurl.com/scikit-tfidf}.
\newblock (Accessed on 04/13/2022).

\bibitem[{Shin et~al.(2020)Shin, Razeghi, Logan~IV, Wallace, and
  Singh}]{shin-etal-2020-autoprompt}
Taylor Shin, Yasaman Razeghi, Robert~L. Logan~IV, Eric Wallace, and Sameer
  Singh. 2020.
\newblock \href {https://doi.org/10.18653/v1/2020.emnlp-main.346}
  {{A}uto{P}rompt: {E}liciting {K}nowledge from {L}anguage {M}odels with
  {A}utomatically {G}enerated {P}rompts}.
\newblock In \emph{Proceedings of the 2020 Conference on Empirical Methods in
  Natural Language Processing (EMNLP)}, pages 4222--4235, Online. Association
  for Computational Linguistics.

\bibitem[{Shoeybi et~al.(2019)Shoeybi, Patwary, Puri, LeGresley, Casper, and
  Catanzaro}]{shoeybi2019megatron}
Mohammad Shoeybi, Mostofa Patwary, Raul Puri, Patrick LeGresley, Jared Casper,
  and Bryan Catanzaro. 2019.
\newblock Megatron-lm: Training multi-billion parameter language models using
  model parallelism.
\newblock \emph{arXiv preprint arXiv:1909.08053}.

\bibitem[{Shutterstock(2013)}]{GitHubLD98:online}
Shutterstock. 2013.
\newblock List of dirty, naughty, obscene, and otherwise bad words.
\newblock
  \url{https://github.com/LDNOOBW/List-of-Dirty-Naughty-Obscene-and-Otherwise-Bad-Words}.
\newblock (Accessed on 12/07/2021).

\bibitem[{Silva et~al.(2020)Silva, Ferreira, Ramos, and
  Paraboni}]{silva2020data}
Samuel Caetano~da Silva, Thiago~Castro Ferreira, Ricelli Moreira~Silva Ramos,
  and Ivandr{\'e} Paraboni. 2020.
\newblock Data-driven and psycholinguistics-motivated approaches to hate speech
  detection.
\newblock \emph{Computaci{\'o}n y Sistemas}, 24(3):1179--1188.

\bibitem[{Simonite(2021)}]{AIandthe49:online}
Tom Simonite. 2021.
\newblock Ai and the list of dirty, naughty, obscene, and otherwise bad words |
  wired.
\newblock
  \url{https://www.wired.com/story/ai-list-dirty-naughty-obscene-bad-words/}.
\newblock (Accessed on 12/07/2021).

\bibitem[{Smith et~al.(2022)Smith, Patwary, Norick, LeGresley, Rajbhandari,
  Casper, Liu, Prabhumoye, Zerveas, Korthikanti et~al.}]{Patwary2021using}
Shaden Smith, Mostofa Patwary, Brandon Norick, Patrick LeGresley, Samyam
  Rajbhandari, Jared Casper, Zhun Liu, Shrimai Prabhumoye, George Zerveas,
  Vijay Korthikanti, et~al. 2022.
\newblock Using deepspeed and megatron to train megatron-turing nlg 530b, a
  large-scale generative language model.
\newblock \emph{arXiv preprint arXiv:2201.11990}.

\bibitem[{S{\o}gaard and Goldberg(2016)}]{sogaard-goldberg-2016-deep}
Anders S{\o}gaard and Yoav Goldberg. 2016.
\newblock \href {https://doi.org/10.18653/v1/P16-2038} {Deep multi-task
  learning with low level tasks supervised at lower layers}.
\newblock In \emph{Proceedings of the 54th Annual Meeting of the Association
  for Computational Linguistics (Volume 2: Short Papers)}, pages 231--235,
  Berlin, Germany. Association for Computational Linguistics.

\bibitem[{Song et~al.(2022)Song, Kim, Park, Shin, and Lee}]{song2022learning}
Hwanjun Song, Minseok Kim, Dongmin Park, Yooju Shin, and Jae-Gil Lee. 2022.
\newblock Learning from noisy labels with deep neural networks: A survey.
\newblock \emph{IEEE Transactions on Neural Networks and Learning Systems}.

\bibitem[{Sood et~al.(2012)Sood, Churchill, and Antin}]{sood2012automatic}
Sara~Owsley Sood, Elizabeth~F Churchill, and Judd Antin. 2012.
\newblock Automatic identification of personal insults on social news sites.
\newblock \emph{Journal of the American Society for Information Science and
  Technology}, 63(2):270--285.

\bibitem[{Spacy(2022)}]{Spacy-POS:online}
POS-tagger Spacy. 2022.
\newblock Part-of-speech-tagging.
\newblock \url{https://spacy.io/usage/linguistic-features#pos-tagging}.
\newblock (Accessed on 04/13/2022).

\bibitem[{Strubell et~al.(2019)Strubell, Ganesh, and
  McCallum}]{strubell2019energy}
Emma Strubell, Ananya Ganesh, and Andrew McCallum. 2019.
\newblock Energy and policy considerations for deep learning in nlp.
\newblock In \emph{Proceedings of the 57th Annual Meeting of the Association
  for Computational Linguistics}, pages 3645--3650.

\bibitem[{SurgeAI(2021)}]{GitHubsu77:online}
SurgeAI. 2021.
\newblock Github - surge-ai/profanity: The world's largest profanity list.
\newblock \url{https://github.com/surge-ai/profanity}.
\newblock (Accessed on 12/07/2021).

\bibitem[{Ullmann and Tomalin(2020)}]{ullmann2020quarantining}
Stefanie Ullmann and Marcus Tomalin. 2020.
\newblock Quarantining online hate speech: technical and ethical perspectives.
\newblock \emph{Ethics and Information Technology}, 22(1):69--80.

\bibitem[{von Ahn(2021)}]{UsefulRe32:online}
Luis von Ahn. 2021.
\newblock Offensive/profane word list.
\newblock \url{https://www.cs.cmu.edu/~biglou/resources/}.
\newblock (Accessed on 12/07/2021).

\bibitem[{Waseem et~al.(2017)Waseem, Davidson, Warmsley, and
  Weber}]{waseem2017understanding}
Zeerak Waseem, Thomas Davidson, Dana Warmsley, and Ingmar Weber. 2017.
\newblock Understanding abuse: A typology of abusive language detection
  subtasks.
\newblock \emph{arXiv preprint arXiv:1705.09899}.

\bibitem[{Waseem and Hovy(2016)}]{waseem-hovy-2016-hateful}
Zeerak Waseem and Dirk Hovy. 2016.
\newblock \href {https://doi.org/10.18653/v1/N16-2013} {Hateful symbols or
  hateful people? predictive features for hate speech detection on {T}witter}.
\newblock In \emph{Proceedings of the {NAACL} Student Research Workshop}, pages
  88--93, San Diego, California. Association for Computational Linguistics.

\bibitem[{Weidinger et~al.(2021)Weidinger, Mellor, Rauh, Griffin, Uesato,
  Huang, Cheng, Glaese, Balle, Kasirzadeh et~al.}]{weidinger2021ethical}
Laura Weidinger, John Mellor, Maribeth Rauh, Conor Griffin, Jonathan Uesato,
  Po-Sen Huang, Myra Cheng, Mia Glaese, Borja Balle, Atoosa Kasirzadeh, et~al.
  2021.
\newblock Ethical and social risks of harm from language models.
\newblock \emph{arXiv preprint arXiv:2112.04359}.

\bibitem[{Zampieri et~al.(2019)Zampieri, Malmasi, Nakov, Rosenthal, Farra, and
  Kumar}]{zampieri-etal-2019-predicting}
Marcos Zampieri, Shervin Malmasi, Preslav Nakov, Sara Rosenthal, Noura Farra,
  and Ritesh Kumar. 2019.
\newblock \href {https://doi.org/10.18653/v1/N19-1144} {Predicting the type and
  target of offensive posts in social media}.
\newblock In \emph{Proceedings of the 2019 Conference of the North {A}merican
  Chapter of the Association for Computational Linguistics: Human Language
  Technologies, Volume 1 (Long and Short Papers)}, pages 1415--1420,
  Minneapolis, Minnesota. Association for Computational Linguistics.

\bibitem[{Zhang et~al.(2018)Zhang, Robinson, and Tepper}]{zhang2018detecting}
Ziqi Zhang, David Robinson, and Jonathan Tepper. 2018.
\newblock Detecting hate speech on twitter using a convolution-gru based deep
  neural network.
\newblock In \emph{European semantic web conference}, pages 745--760. Springer.

\bibitem[{Zhong et~al.(2021)Zhong, Friedman, and
  Chen}]{zhong-etal-2021-factual}
Zexuan Zhong, Dan Friedman, and Danqi Chen. 2021.
\newblock \href {https://doi.org/10.18653/v1/2021.naacl-main.398} {Factual
  probing is [{MASK}]: Learning vs. learning to recall}.
\newblock In \emph{Proceedings of the 2021 Conference of the North American
  Chapter of the Association for Computational Linguistics: Human Language
  Technologies}, pages 5017--5033, Online. Association for Computational
  Linguistics.

\bibitem[{Zhou et~al.(2020)Zhou, Zhang, Cui, and
  Huang}]{zhou-etal-2020-evaluating}
Xuhui Zhou, Yue Zhang, Leyang Cui, and Dandan Huang. 2020.
\newblock Evaluating commonsense in pre-trained language models.
\newblock In \emph{Proceedings of the AAAI Conference on Artificial
  Intelligence}, volume~34, pages 9733--9740.

\end{thebibliography}
\bibliographystyle{acl_natbib}

\begin{table*}[t]
\centering
\small{
\begin{tabular}{@{}p{0.5in}@{\hskip 0.2in} p{4.0in}@{\hskip 0.2in} p{1.3in}@{}}
\textbf{Task} & \textbf{Definition} & \textbf{Tokens}\\
\toprule
Offensive & Does this post contain offensive language? & Yes, No \\
%& Could this post be considered offensive, disrespectful, or toxic to anyone/someone? \\
Intent & Does this post contain intentional insults?  & Yes, No \\
%& Does this post contain offensive jokes, insults, personal attacks, profanity, or aggression? \\
Lewd & Does this post contain sexual content? & Yes, No \\
%& Does this post contain or allude to sexual content/acts (This includes euphemism, double entendre, etc)? \\
Group & Does this post contain offense to a group? & Yes, No  \\
%& Could this post be offensive to a group (minority, identity-based, etc.) \\
WHO & Who is referred to/targeted by this post - body, culture, disabled, gender, race, social, victim? & body, culture, disabled, gender, race, social, victim\\
HOF & Does this post contain any form of non-acceptable language such as hate speech, offensiveness, aggression, profanity?  & Yes, No \\
HOP & This post contains & hate, offense, profanity\\
%& What does does this post contain - hate, offense, profanity? \\
Target & Does this post contain an insult/threat to an individual, group, or others?  & Yes, No \\
\bottomrule
\end{tabular}
}
%\vspace{-0.5em}
\caption{Definitions of bias used for each task. We formulate the definitions based on labeling instruction from the SBIC and HASOC datasets. For binary definitions we use a template in the form of: \textit{Does this post contain <attribute>?} }
%\vspace{-1.0em}
\label{tab:bias_defn}
\end{table*}

\newpage
\newpage
\appendix

\section{Appendix}
\label{sec:appendix}

\subsection{Bias Definitions}
\label{sec:bias_def}

The bias definitions used in all experiments are mentioned in Table~\ref{tab:bias_defn}.
We formulate the definitions for binary tasks based on labeling instruction from the datasets.
We use a template for the definition in the form of: \textit{Does this post contain \{attribute\}?}.
We also mention the tokens for which we calculate classification probability in Table~\ref{tab:bias_defn}.

\subsection{Keyword analysis}
\label{sec:keyword_analysis}

We study the percentage of keywords present in the textual posts of the test set for each binary task and their correlation to labels.
We use a superset of keyword ($S$) containing 3173 keywords which is a union of keywords from \cite{UsefulRe32:online, GitHubsu77:online, GitHubLD98:online}.
We calculate the overlap of keywords ($S$) with a textual post $\mathbf{Q}$.
Specifically, we check its correlation with the \textit{pos} and \textit{neg} labels for the binary tasks i.e percentage of positive textual posts $p$ that have at least one keyword from $S$ and percentage of negative textual posts $n$ that have at least one keyword from $S$. 
The ratio is calculated as $p/n$ and tells us the rate at which positive posts have a higher/lower overlap with keywords compared to the negative posts.
This analysis along with label correlation is presented in Table~\ref{tab:keyword_sbic}.
We can see that the \emph{Target} task from HASOC has the highest ratio indicating that it has higher overlap of keywords with the positive labeled examples as compared to the negative labeled examples leading to a hard to beat heuristic.

\begin{table}[t]
\centering
\small{
\begin{tabular}{@{}l r r r@{}}
\textbf{Task} & \textbf{pos} & \textbf{neg} & \textbf{ratio} \\
\toprule
Offensive (57.8\% pos) & 85.20 & 74.51 & 1.14 \\
Intent (53.1\% pos) & 84.75 & 76.08 & 1.11 \\
Lewd (9.6\% pos) & 93.35 & 79.34 & 1.18 \\
Group (41.1 \% pos) & 84.89 & 77.76 & 1.09 \\
HOF (25.0\% pos) & 88.54 & 76.07 & 1.16 \\
Target (85.1\% pos) & 88.57 & 15.51 & 5.71 \\
\bottomrule
\end{tabular}
}
%\vspace{-0.5em}
\caption{Keyword analysis for the SBIC tasks, based on union of keywords from \cite{UsefulRe32:online, GitHubsu77:online, GitHubLD98:online}. The percentages show the overlap of keywords with the \textit{pos} and \textit{neg} labels for the binary tasks. We can see that the \emph{Target} task from HASOC has the highest overlap of keywords with gold labels leading to a hard to beat heuristic. We show the percentage of positive samples in the test set in brackets.}
%\vspace{-1.0em}
\label{tab:keyword_sbic}
\end{table}

\subsection{Qualitative Analysis}
\label{sec:qualitative_analysis}

Qualitative analysis as described in section \Sref{sec:analysis} is shown in Table~\ref{tab:qualitative_analysis}. 
We show the representative textual posts for each category (Warning: the examples shown contain content that maybe offensive or upsetting).

\begin{table*}[!t]
\centering
%\resizebox{\textwidth}{!}{
\small{
\begin{tabular}{@{}p{0.11in} p{2.2in}@{\hskip 0.2in} p{0.3in}@{\hskip 0.1in} p{2.8in}@{\hskip 0.2in}  p{0.2in}@{}}
\textbf{Id} & \multicolumn{1}{c}{\textbf{Error Class}}{\hskip 0.2in} & \textbf{\%}{\hskip 0.1in} & \multicolumn{1}{c}{\textbf{Textual Post}}{\hskip 0.2in} & \textbf{O.}\\

\toprule
\multicolumn{5}{c}{Cases where MT-NLG makes correct prediction and TF-IDF is wrong} \\
\toprule
1 & Post is offensive and does not contains offensive keywords &  38.33 & 
What's black, white and rolls around on the beach? A black man and a seagull fighting over a chip.. & Y \\
2 & Post is not offensive and does not contain offensive keywords & 38.33 & 
Bush: 'it's hard to unify the country when the news media is so split up. Remember when the news media mattered a lot more'. & N \\
3 & Post is offensive and contains offensive keywords & 10.00 & 
Y'RE ALL ANNOYING B**CH YOU'RE NOT MY CLASSMATES ANYMORE.
%I had to miss the opening night of Black Panther with my black friends. Safe to say, that klan meeting was worth leaving them hanging. 
& Y \\
4 & Post is not offensive but contains keywords & 8.33 &
RT @markiplier: In the car with my frands going F***ING HYPERSPEED. & N \\
5 & Mislabeled post & 5.00 &
omg twitter you are killing me with dumb this morning. & N\\

\toprule
\multicolumn{5}{c}{Cases where MT-NLG makes wrong prediction and TF-IDF is correct} \\
\toprule
6 & Post is not offensive and does not contain offensive keywords & 31.67 & 
The answer is yes I can fit 2 more people in this shirt therefor nobody will know I'm not wearing a bra. & N \\
7 & Post is offensive and does not contain offensive keywords & 26.67 & 
I know I keep hoping I wake up one day and this multiracial hell Canada was all just a terrible nightmare, but it never happens. & Y \\
8 & Post is offensive and contains offensive keywords & 21.67 & 
number of femoids who would rather die than spend more than 5 seconds with an incel. & Y \\
9 & Post is not offensive but contains keywords & 15.00 & 
AHHH karmas a b**ch. & N\\
11 & Mislabeled post & 5.00 & Are you a bummer tied to a tree? No?? BUMMER ON THE LOOSE!! & N \\
\bottomrule
\end{tabular}
}
%\vspace{-0.5em}
\caption{Analysis of the offensive task from \cite{sap-etal-2020-social}. \textbf{Textual Post} indicates grounded information. \textbf{O.} indicates the ground truth label for the category.}
%\vspace{-1.0em}
\label{tab:qualitative_analysis}
\end{table*}

\subsection{Details on Finetuned Models}
\label{sec:finetune_details}

We provide details of the \textbf{SC} and \textbf{HS} models described in \Sref{sec:results}.
The \textbf{SC} model~\cite{sap-etal-2020-social} was trained to predict the entire social bias frame given the text of the post i.e given the post the GPT-2 model was finetuned to generate the token corresponding to each of the five classes in the original set ($w_{[\text{lewd}]}, w_{[\text{offensive}]}, w_{[\text{intent}]}, w_{[\text{group}]}, w_{[\text{in-group}]}$). 
Note that we don't present results for the \textit{in-group} classification task.

We report the results of best performing systems in English in the HASOC track at FIRE 2019: Hate Speech and Offensive Content Identification in Indo-European Languages~\cite{mandl-etal-2019-hasoc}. 
We report the result of YNU\_wb team for \textbf{\textit{HOF}}task, and 3Idiots team for \textbf{\textit{HOP}} and \textbf{\textit{target}} task.

\begin{table}[t]
\centering
\small{
\begin{tabular}{@{}l l r r@{}}
\textbf{Model} & \textbf{Sampling} & \multicolumn{2}{c}{\textbf{Intent}}\\
\toprule
 & & AUC & F1 \\
\cmidrule{3-4}
SC & - & - & 78.60 \\
\cmidrule{3-4}
\sbert & \sbert & 79.36 & 82.36 \\
\mOne & \sbert &  64.47 & 68.05 \\
\mEight & \sbert & 68.95 & 75.42 \\
\ours & \sbert & 77.59 & 80.00 \\
\bottomrule
\end{tabular}
}
%\vspace{-0.5em}
\caption{Results for the $32$-shot prompting on intent classification task from SBIC dataset \Sref{sec:datasets} on \sbert{} baseline and three language models which use shots sampled by \sbert{}.}
%\vspace{-1.0em}
\label{tab:sbert_res}
\end{table}

\section{Analysis on Sentence-BERT}
\label{sec:sbert}

In this section, we show additional experiments using Sentence-BERT (\sbert{}) sampling technique~\cite{reimers-2019-sentence-bert} to select class-balanced shots for $k$-shot classification. 
Although, the experiments in Table~\ref{tab:sbert_res} show that \sbert{} can perform better for the bias tasks explored in this paper, results in Table~\ref{tab:data_strat} and Table~\ref{tab:anli_res} show this technique does not generalize to other tasks. 
We show in Section~\Sref{sec:sbert_add_analysis} that these sampling techniques take advantage of the keywords that are used only in one specific contexts.

We use \sbert{} to encode the post  $\mathbf{Q}$ and the posts from $\mathcal{D}$ into a common embedding space.\footnote{\url{https://huggingface.co/sentence-transformers/all-MiniLM-L12-v2}}
We then select $k/|\mathbf{C}|$ posts with the highest cosine similarity score from each class.
These shots are provided as context to language model $\mathcal{M}$.

For the \sbert{} baseline, we consider the class that has the highest average score as the prediction.

\paragraph{Results}
The results for $32$-shot intent classification task on the \sbert{} baseline and the LMs using \sbert{} sampling technique are shown in Table~\ref{tab:sbert_res}.
We observe that \sbert{} performs better than all the language models in intent classification task.

\begin{table}[t]
\centering
\small{
\begin{tabular}{@{}l l r r@{}}
\textbf{Model} & \textbf{Sampling} & \multicolumn{2}{c}{\textbf{Intent}}\\
\toprule
 & & AUC & F1 \\
\cmidrule{3-4}
\sbert & \sbert-strat & 23.51 & 14.54  \\
\mOne & \sbert-strat & 58.16 & 60.32 \\
\mEight & \sbert-strat & 64.52 & 71.45 \\
\ours & \sbert-strat & 74.15 & 76.43  \\
\bottomrule
\end{tabular}
}
%\vspace{-0.5em}
\caption{Results for the $32$-shot prompting on intent classification task on data stratified \sbert{} baseline and three language models which use shots sampled by data stratified \sbert{}.}
%\vspace{-1.0em}
\label{tab:data_strat}
\end{table}

\paragraph{Data Stratification}

The SBIC dataset exhibits strong correlation between semantic similarity (captured by \sbert{}) and target class. This creates a very strong heuristic baseline, that is hard to beat even for fine-tuned model (see Table \ref{tab:sbert_res}). To remove the impact of such simple heuristic and investigate whether the model is capable of reasoning on the shot content, we implemented a stratified randomization for balanced shot selection. The goal of the stratification is to select equal number of shots for each class, such that the mean difference in cosine similarity of the class-balanced shots is minimized. To accomplish this, we implemented a binning-based stratification of shot-query cosine similarity scores \cite{lohr2021sampling}.

After \sbert{} embeddings and cosine similarities are calculated for all the train set, the stratified shot selection follows the following steps:
1) shots are ordered based on cosine similarity with the query text,
2) shots are allocated into bins based on their cosine similarity score using bin thresholds calculated using numpy implementation of histogram based binning \footnote{`auto' binning - takes the maximum of \emph{Sturges} and \emph{Freedman Diaconis} estimators - \url{https://numpy.org/doc/stable/reference/generated/numpy.histogram_bin_edges.html}}, 
3) starting from the bin with the highest similarity and moving downwards, a maximum number of class-balanced shots from each bin is selected, and 
4) step 3 is repeated until the total desired number of shots are selected.

As a result of this process, the average difference in mean cosine similarity between the shots for two classes is $0.0029$ ($SD$=$0.0020$). %Without the stratification this difference amounts to MEAN (SD=)
The summary of shot stratification impact can be seen in Table \ref{tab:data_strat}.
Using stratified class-balanced shots, removes the impact of simple \sbert{} heuristic reducing AUC from $79.36$ to $23.52$ and F1 from $82.36$ to $14.54$, but at the same time has limited impact on the performance of the \ours{} model reducing AUC by $4.4\%$ ($77.59$ to $74.15$) and F1 by $4.5\%$ ($80.00$ to $76.43$). 

The results in Table~\ref{tab:data_strat} illustrates the strong correlation between semantic similarity captured by \sbert{} and target class.
We would like to note that this correlation is specific to bias datasets since often times they are collected using certain referential terms and keywords.
To further bolster our hypothesis we present additional analysis of correlation of keyword in the shots selected by \sbert{} and the label, as well as an analysis on ANLI task (see \Sref{sec:sbert_add_analysis}).

\subsection{Analysis}
\label{sec:sbert_add_analysis}

\paragraph{Keyword overlap}
We performed additional in-depth analysis to better understand the strong performance of \sbert{} heuristic on the SBIC dataset. Through qualitative exploration of the ``Intent'' task we observed an overlap of the key referential terms present in test set queries, such as \emph{``clinton''}, \emph{``p***phile''} predominantly with shots for one label, but not the other. This could suggest that in the SBIC dataset, the presence of such terms is sufficient to decide on the label. To verify this observation at scale, we lemmatized the test queries extracting only NOUNS and PRONOUNS (via part-of-speech tagging \cite{Spacy-POS:online}). This process extracted referential terms such as: ``clinton'', ``p***phile'', ``b**ch'', ``lord'', etc. We further counted the frequency of these terms in shots picked from the train set with the same and opposite label (based on gold). For example, for post \textit{`Alex Jones \& Mike Cernovich: ``It's crazy how the P***philes \ldots all LOOK LIKE P***PHILES" \ldots \# Truth'} the extracted terms are `alex', `cernovich', `jones', `mike', `p***phile', `truth'. The frequency of these terms in same labeled shots resolves to: `p***phile': 12, `cernovich': 1, `alex': ', `jones': 1, while for the opposite label shots these terms as much less frequent: `p***phile': 6, `cernovich': 1, `truth': 2. From this we can see that e.g., the term ``p***phile'' is much more frequent in the shots with the same label, which suggests it is used in only one context.

At the scale of the entire test set, we quantified that the same label shots have on average 11.36 terms overlapping with the test posts, while for the opposite label shots this overlap amounts to only 7.76. Furthermore the ratio of keyword overlap between same and opposite label shots is 1.97 on average, meaning that there are almost 2x as many keywords overlapping with test post for the same labeled shots compared to opposite labeled shots.

This analysis suggests that crucial referential terms present in test posts, which, in essence, could be used in a biased or unbiased context (i.e., even term ``p***phile'' could be used in challenging contexts, such as a policy announcement or criticisms) are used in only one context. Such strong association of terms and context permits the mere presence or absence of such referential terms to be sufficient to decide on the label, which in turn drives high performance of simple \sbert{} heuristic.

\paragraph{ANLI Results}

To further investigate our hypothesis that bias datasets have the property of using certain words or phrases only in biased or non-biased context, we test our approach on another task - the ANLI task.

The ANLI~\cite{nie2020adversarial} dataset is an adversarially mined natural language inference (NLI) dataset that aims to create a difficult set of NLI problems. 
It has 3 iterative rounds of data collection marked as ANLI-R1, ANLI-2 and ANLI-R3.
Following~\cite{Patwary2021using}, we rephrase the NLI problem into a question-answering format where each example is structured as ``<premise>Question:<hypothesis>. True, False or Neither?Answer:''.
This prompt is given to the language model and we calculate the probability of tokens ``True'', ``False'' and ``Neither''.
The token with the highest likelihood assigned by the model is considered as the model prediction.

In case of $k$-shot classification, we use \sbert{} to select class-balanced shots which are most similar to the query.
The results of accuracy scores for the three rounds of ANLI datasets are shown in Table~\ref{tab:anli_res} with $k=\{48, 24, 9\}$.
We observe that for all values of $k$, \ours{} performs significantly better than the \sbert{} baseline.
ANLI datasets were collected with the goal of having diverse set of contexts from various domains.
They contain text extracted from Wikipedia, News (extracted from Common Crawl), fiction (extracted from StoryCloze and CBT), formal spoken text (excerpted from court and presidential debate transcripts in the Manually Annotated Sub-Corpus (MASC) of the Open American National Corpus3) and causal or procedural text, which describes sequences of events or actions, extracted from WikiHow (see Section 2.5 in \citep{nie2020adversarial}).
Hence, the dataset does not contain samples that use certain phrases only in one context or uses certain words/phrases only for one label.
Hence, this further solidifies our argument that our approach is generally applicable but struggles to perform better than heuristic-based \sbert{} baseline when the dataset is skewed.

\begin{table}[t]
\centering
\small{
\begin{tabular}{@{}l l r r r@{}}
\textbf{Model} & $\mathbf{k}$ & \textbf{ANLI-R1} & \textbf{ANLI-R2} & \textbf{ANLI-R3} \\
\toprule
\sbert & 48 & 28.10 & 30.80 & 30.16 \\
\ours & 48 & 40.20 & 44.40 & 49.08 \\
\sbert & 24 & 27.10 & 31.90 & 30.75 \\
\ours & 24 & 43.00 & 45.60 & 50.42 \\
\sbert & 9 & 28.50 & 38.50 & 35.08 \\
\ours & 9 & 45.00 & 44.90 & 51.33 \\
\bottomrule
\end{tabular}
}
%\vspace{-0.5em}
\caption{Accuracy for the $k$-shot prompting on three different natural language inference datasets on the \sbert{} baseline and \ours{} with $k=\{48, 24, 9\}$ }
%\vspace{-1.0em}
\label{tab:anli_res}
\end{table}

\section{Ethical Considerations}
\label{sec:ethics}
%Intended use
The intended use of the proposed instruction-based detection techniques is to aid the identification of different forms of bias either in human or AI produced textual content. 
One potential applications can be to assist moderation of content in settings such as social media or in end-user AI applications involving language generation such as conversational agents. 
It can also be used to add safety features to the generation outputs of large LMs as well as it can assist in building future large LMs that are not biased by identifying biases in pretraining data.
Given the limitations of our proposed method, it should likely not be used as a sole measure for detecting bias, but we believe it could serve as a low-effort initial filter and feedback mechanism.
Furthermore, we see its use as potential means for promoting positive online interaction and higher community standards \cite{does2011thou}.
% should not be used as a sole measure
% promotion of positive online interactions 
% community standards counter-speech
% imperfect performance that may necessitate supplementing with further techniques

Our proposed method, can unfortunately be misused intentionally or unintentionally \cite{weidinger2021ethical}. 
We specifically see the dangers of using our approach for censorship \cite{ullmann2020quarantining} or limiting expression in specific settings where violent or lewd language might be intended (e.g., crime fiction). 
Furthermore, there is danger of misappropriating our approach to introduce racially targeted censorship based on dialect \cite{sap2019risk}. 
Despite certain degree of robustness and our additional experiments (e.g., with flipped labels), our method still relies on supervised labeling of few-shots provided as context, and as such can be affected by the imperfections of the labeling, such as racial bias in existing labeled datasets \cite{davidson2019racial}.
As shown in recent work, human annotation process can be a source of bias in itself \cite{hovy2021five} and labelers may even be selected by malicious actors to introduce biased interpretation on purpose \cite{flekova2016analyzing}.

\end{document}